\documentclass[]{bytedance_seed}

\usepackage[toc,page,header]{appendix}

\usepackage{minitoc}
\usepackage{url}
\usepackage[utf8]{inputenc} 
\usepackage[T1]{fontenc}    
\usepackage{hyperref}       
\usepackage{url}            
\usepackage{booktabs}       
\usepackage{amsfonts}       
\usepackage{nicefrac}       
\usepackage{microtype}      
\usepackage{xcolor}         
\usepackage{makecell}
\usepackage{graphicx}
\usepackage{amsmath}
\usepackage{multirow}
\usepackage{wrapfig}
\usepackage{enumitem}
\usepackage{wrapfig}
\usepackage{float}
\usepackage{graphicx}
\usepackage{caption}
\usepackage{color}
\usepackage{colortbl}
\usepackage{xcolor}
\usepackage{bbding}
\usepackage{pifont}
\usepackage{tabularx} 
\usepackage{booktabs}
\usepackage{algorithmic}
\usepackage{algorithm}
\usepackage{natbib}   

\usepackage[most]{tcolorbox}
\usepackage{xcolor}
\usepackage{fontawesome5}
\usepackage{wrapfig}
\usepackage{tabularx}
\usepackage{adjustbox}
\usepackage{array}
\usepackage{xcolor}
\usepackage{graphicx}

\newcommand{\cmark}{\textcolor{green!55!black}{\ding{51}}}
\newcommand{\xmark}{\textcolor{red!70!black}{\ding{55}}}
\newcommand{\pmark}{\textcolor{orange!85!black}{$\triangle$}}

\newcommand{\PKLabel}[1]{%
  {\normalfont\bfseries\color{blue!85!black}[#1]}%
}

\newcommand{\PK}[2]{%
  \PKLabel{#1}~{\bfseries\color{blue!50!black}#2}%
}

\definecolor{rqcyan}{HTML}{5AC5D0}

\newtcolorbox{researchquestionbox}{
  enhanced,
  colback=rqcyan!6!white,
  colframe=rqcyan,
  boxrule=1.0pt,
  arc=4mm,
  left=2mm,
  right=2mm,
  top=1.5mm,
  bottom=1.5mm,
  before skip=0.6em,
  after skip=0.8em
}

\title{Morphing into Hybrid Attention Models}

\author[1,2,*]{Disen Lan}
\author[2]{Jianbin Zheng}
\author[2]{Yuxi Ren}
\author[2]{Xin Xia}
\author[2]{Xuanda Wang}
\author[2]{\\Xuefeng Xiao}
\author[1,\dagger]{Xipeng Qiu}
\author[3,\dagger]{Yu Cheng}

\affiliation[1]{Fudan University}
\affiliation[2]{ByteDance Seed}
\affiliation[3]{The Chinese University of Hong Kong}

\contribution[*]{Work done at ByteDance Seed}
\contribution[\dagger]{Corresponding authors}

\abstract{
Hybrid attention models improve long-context efficiency by retaining only a subset of full-attention layers and replacing the remaining layers with linear attention. However, the effectiveness of Transformer-to-hybrid conversion critically depends on which layers preserve full attention. Existing hybrid layer selection methods typically rely on heuristic strategies such as fixed placement patterns or layerwise scoring, implicitly treating layer importance as isolated and overlooking the interdependent layer effect under a global hybrid configuration. In this work, we formulate hybrid layer selection as a budget-constrained subset optimization problem. We further propose \textbf{FlashMorph} (\underline{\textbf{F}}ast \underline{\textbf{LA}}yer \underline{\textbf{S}}election for \underline{\textbf{H}}ybrid \underline{\textbf{MORPH}}ing), an effective, efficient and scalable layer selection method for Transformer-to-hybrid conversion. FlashMorph first constructs a morphable model by equipping each full-attention layer with a converted linear-attention branch. It then freezes all model weights and jointly optimizes layerwise gates on synthetic long-context retrieval data, with a linearization regularization that encourages the model to rely on linear attention for efficiency. The learned gates are discretized under a preset full-attention budget to instantiate the hybrid architecture, followed by standard logits distillation and long-context finetuning. Extensive experiments show that FlashMorph discovers more effective hybrid configurations, preserves strong long-context recall and general benchmark performance while substantially reducing layer selection cost compared with existing layer selection methods, demonstrating its effectiveness, efficiency, and scalability.
}

\date{\today}

\correspondence{Xipeng Qiu at \email{xpqiu@fudan.edu.cn}, Yu Cheng at \email{chengyu@cse.cuhk.edu.hk}}
\checkdata[Code]{\url{https://github.com/LanDisen/FlashMorph}}

\begin{document}
\maketitle


\vspace{-2.0em}
\section{Introduction}

The Transformer architecture~\citep{vaswani2017attention} has become the dominant backbone of modern large language models (LLMs), driving substantial progress in sequence modeling and complex  reasoning~\citep{grattafiori2024llama,cai2024internlm2,liu2024deepseek,guo2025deepseek,yang2025qwen3}. Nevertheless, its reliance on softmax attention introduces a fundamental efficiency bottleneck: increasing sequence length leads to quadratic growth in attention computation and linear growth in the Key-Value (KV) cache required for autoregressive inference~\citep{kwon2023efficient,sun2025speed}. These limitations have motivated the development of more efficient sequence mixers, such as linear attention~\citep{katharopoulos2020transformers,yang2023gated,qin2024lightning,qin2024hgrn2,yang2024parallelizing,yang2024gated} and state-space models~\citep{gu2023mamba,dao2024transformers,lahoti2026mamba,hu2025comba}, which reduce the computational cost of long-sequence modeling and eliminate the need for a growing KV cache by maintaining fixed-size recurrent states. However, purely linear recurrent architectures are generally less effective than Transformer-based LLMs on long-context and recall-sensitive tasks~\citep{arora2024simple,wen2025rnns}. This performance gap has motivated hybrid attention architectures, which retain full attention in a subset of layers while replacing the remaining layers with efficient linear sequence mixers, achieving a more favorable trade-off between model quality and efficiency~\citep{chen2025minimax,blakeman2025nemotron,zuo2025falcon,qwen_qwen3_coder_next_tech_report,team2025kimi_linear,merrill2026olmo}.

However, training high-quality hybrid LLMs from scratch remains prohibitively expensive. A more practical alternative is \textbf{Transformer-to-hybrid Conversion}, which starts from a pretrained Transformer-based LLM, retains only a small subset of its full-attention layers, and replaces the remaining layers with efficient linear attention through model parameter transfer, distillation and continued finetuning~\citep{kasai2021finetuning,mercat2024linearizing,wang2024mamba,bick2024transformers,zhang2024lolcats,lan2025liger}. By reusing the weights of a strong Transformer model, this paradigm avoids the significant cost of training a hybrid architecture from scratch while preserving much of the original model capability. Nevertheless, its effectiveness critically depends on how the limited budget of retained full attention is allocated across layers.

In principle, given an $L$-layer model and a budget of $K$ retained full-attention layers, identifying the optimal hybrid configuration requires evaluating all $\binom{L}{K}$ possible subsets, which quickly becomes intractable~\citep{li2025distilling}. Existing hybrid layer selection methods can therefore be understood as heuristic, tractable approximations to this combinatorial subset-selection problem. Uniform interleaving avoids search altogether by imposing fixed attention placement rules, but it ignores the pretrained model and the heterogeneous functional roles of different layers. Search-based methods explore layer placements through auxiliary architecture search or supernet training, but they introduce substantial optimization and evaluation overhead~\citep{gu2025jet}. Layerwise methods estimate the marginal utility of each layer by perturbing, replacing, or restoring one layer at a time, and then select layers according to the resulting scores~\citep{yang2026zebra,li2025distilling,chen2026hybrid}. More fundamentally, these heuristic approximations reduce hybrid layer selection to either fixed attention placement rules or isolated layer scoring, implicitly assuming that the effect of a multi-layer hybrid configuration can be inferred from individual layer importance. This assumption overlooks interdependent layer effects and fails to capture the joint effects that emerge when multiple layers are converted together. Moreover, layerwise methods require repeated one-layer perturbation or restoration to estimate layer importance, incurring substantial selection overhead and making them costly to scale. This motivates the central question studied in this work:

\begin{researchquestionbox}
\quad
\begin{minipage}[t]{0.95\linewidth}
\itshape
Can hybrid layer selection be formulated as a budget-constrained joint optimization problem that globally accounts for interdependent layer effect, rather than relying on heuristic approximations such as fixed attention placement rules or isolated layerwise scoring?
\end{minipage}
\end{researchquestionbox}

We argue that hybrid layer selection should be treated as a subset selection problem under a global hybrid configuration. In this setting, the contribution of each layer depends on which other layers are retained or converted: jointly converting layers that support related functions may amplify degradation, whereas retaining layers with overlapping roles may yield diminishing returns. The objective is to identify a subset of full-attention layers whose collective effect best balances model quality and efficiency, while accounting for complementarity and redundancy across layers.

\begin{wraptable}{r}{0.45\linewidth}
\vspace{-1.5em}
\caption{\textbf{Comparison of Hybrid Layer Selection (LS) Methods.} LS cost refers to the required tokens for layer selection. \cmark, \xmark, and \pmark indicate fully supported, not supported, and partially supported, respectively.
}
\small
\setlength{\tabcolsep}{3.5pt}
\renewcommand{\arraystretch}{1.12}
\resizebox{\linewidth}{!}{
\begin{tabular}{lccc}
\toprule
\textbf{LS method}
& \textbf{LS cost $\downarrow$}
& \textbf{Non-isolated}
& \textbf{Optimization-based} \\
\midrule
Uniform & N/A & \xmark & \xmark \\
PostNAS & 50B  & \pmark & \cmark \\
KL-LS & 20B  & \xmark & \cmark \\
HALO & 234M & \xmark & \xmark \\
\textbf{FlashMorph (ours)} & \textbf{20M} & \cmark & \cmark \\
\bottomrule
\end{tabular}
}
\label{tab:intro_comparison}
\end{wraptable}

Motivated by this, we propose \textbf{FlashMorph} (\underline{\textbf{F}}ast \underline{\textbf{LA}}yer \underline{\textbf{S}}election for \underline{\textbf{H}}ybrid \underline{\textbf{MORPH}}ing), an effective, efficient, and scalable layer selection method for Transformer-to-hybrid conversion. FlashMorph first constructs a morphable model, in which each pretrained full-attention layer is equipped with a converted linear-attention branch and can be continuously transformed between full and linear attention by introducing layerwise learnable gates. During layer selection, both the pretrained backbone and the converted linear-attention branches are kept frozen, while only the layerwise gates are optimized on synthetic retrieval data. By jointly optimizing all gates, FlashMorph estimates the relative necessity of retaining full attention under a global hybrid configuration, rather than relying on isolated layer scoring. The optimized gates are then discretized according to a prescribed full-attention budget to instantiate a hybrid architecture, which is subsequently trained with standard distillation and long-context finetuning. As summarized in Table~\ref{tab:intro_comparison}, FlashMorph performs non-isolated layer selection that captures inter-layer complementarity and redundancy  through joint optimization, and requires substantially fewer selection tokens than prior methods, thereby yielding a stronger quality-efficiency trade-off for Transformer-to-hybrid conversion.

To summarize, our contributions are as follows:
\begin{itemize}
\item We formulate hybrid layer selection for Transformer-to-hybrid conversion as a budget-constrained joint optimization problem, moving beyond fixed placement rules and isolated layerwise scoring by accounting for the collective effect of retained full-attention layers.

\item We propose FlashMorph, an effective, efficient, and scalable layer selection method that reformulates hybrid layer selection as a joint optimization procedure. FlashMorph constructs a morphable model by pairing each pretrained full-attention layer with a converted linear-attention branch, keeps both branches frozen during selection, and jointly optimizes layerwise gates on synthetic retrieval data to estimate the necessity of retaining full attention under a global hybrid configuration.

\item  We conduct extensive experiments on Qwen3-series models with multiple linear-attention variants, covering long-context retrieval, commonsense reasoning, and recall-intensive tasks. The results show that FlashMorph improves the quality--efficiency trade-off of Transformer-to-hybrid conversion while substantially reducing layer selection cost, demonstrating its effectiveness, efficiency, and scalability.
\end{itemize}
\vspace{-0.3em}
\section{Preliminaries}

\vspace{-0.3em}
\subsection{Notation}

Let \(\mathbf{X} = [\mathbf{x}_1;\mathbf{x}_2;\dots;\mathbf{x}_T] \in \mathbb{R}^{T \times d}\) be an input sequence with the length of $T$, where \(\mathbf{x}_t \in \mathbb{R}^{1 \times d}\) is the token representation at position \(t\) with the dimension of $d$. For simplicity, we omit multi-head notation and write all equations for a single attention head. The query, key, and value vectors are computed as

\vspace{-0.8em}
\begin{equation}
\label{eq:qkv}
    \mathbf{q}_t = \mathbf{x}_t \mathbf{W}_Q,\quad
    \mathbf{k}_t = \mathbf{x}_t \mathbf{W}_K,\quad
    \mathbf{v}_t = \mathbf{x}_t \mathbf{W}_V,
\end{equation}

where \(\mathbf{q}_t,\mathbf{k}_t,\mathbf{v}_t \in \mathbb{R}^{1 \times d}\), and 
\(\mathbf{W}_Q,\mathbf{W}_K,\mathbf{W}_V \in \mathbb{R}^{d \times d}\) are learnable projection matrices.

\vspace{-0.3em}
\subsection{Full Attention}

Full (softmax) attention computes the output at each position by comparing the current query with all previous keys and taking a weighted sum over the corresponding values. Under causal masking, the attention output at position \(t\) is

\vspace{-1.6em}
\begin{equation}
\label{eq:full_attention}
    \mathbf{o}_t
    =
    \sum_{i=1}^{t}
    \alpha_{t,i}\mathbf{v}_i \mathbf{W}_O,
    \quad
    \alpha_{t,i}
    =
    \frac{
    \exp\left(\mathbf{q}_t \mathbf{k}_i^{\top} / \sqrt{d}\right)
    }{
    \sum_{j=1}^{t}
    \exp\left(\mathbf{q}_t \mathbf{k}_j^{\top} / \sqrt{d}\right)
    },
\end{equation}

where $\mathbf{W}_O \in \mathbb{R}^{d \times d}$ is the output projection. 
Because each query attends to all previous keys, full attention explicitly models pairwise token interactions. This gives it strong capacity for precise matching and long-range dependency modeling, but also leads to $\mathcal{O}(T^2)$ computation over the sequence length and KV cache whose memory grows linearly with the context length during autoregressive inference.

\vspace{-0.3em}
\subsection{Linear Attention}

Linear attention replaces the softmax kernel with a formulation that can be accumulated recurrently. 
Let $\phi(\cdot)$ be a feature map applied to queries and keys, a generic kernelized linear attention output can be written as

\vspace{-0.3em}
\begin{equation}
\label{eq:linear_attention}
    \mathbf{o}_t
    =
    \frac{
    \phi(\mathbf{q}_t)
    \sum_{i=1}^{t}
    \phi(\mathbf{k}_i)^{\top}\mathbf{v}_i
    }{
    \phi(\mathbf{q}_t)
    \sum_{i=1}^{t}
    \phi(\mathbf{k}_i)^{\top}
    } \mathbf{W}_O .
\end{equation}

In many modern linear attention variants, the normalization term is omitted, and the key-value statistics are maintained as a recurrent memory state. After absorbing the feature map into \(\mathbf{q}_t\) and \(\mathbf{k}_t\), a broad class of linear attention sequence mixers can be expressed as RNN style

\begin{equation}
\label{eq:linear_state_update}
\begin{aligned}
    \mathbf{S}_t 
    &= 
    \mathbf{A}_t \mathbf{S}_{t-1}
    +
    \mathbf{k}_t^{\top}\mathbf{v}_t, \\
    \mathbf{o}_t 
    &= 
    \mathbf{q}_t \mathbf{S}_t \mathbf{W}_O ,
\end{aligned}
\end{equation}

where $\mathbf{S}_t \in \mathbb{R}^{d \times d}$ is the recurrent memory state and $\mathbf{A}_t \in \mathbb{R}^{d \times d}$ is a state-transition or decay matrix. Unlike full attention, the recurrent formulation avoids storing all previous keys and values, enabling $\mathcal{O}(T)$ linear-time sequence processing and constant-size state caching during autoregressive decoding.

\vspace{-0.3em}
\subsection{Hybrid Attention}

While linear attention is efficient, purely linear models may lose part of the long-range dependency modeling and retrieval ability of full attention. Hybrid attention mitigates this trade-off by retaining full attention in a subset of layers and replacing the remaining layers with linear attention. For an $L$-layer model, let $\mathcal{I}_{\mathrm{full}} \subseteq \{1,\dots,L\}$ denote the set of layers that use full attention. The sequence mixer in layer $l$ is defined as

\begin{equation}
\label{eq:hybrid_mixer}
\operatorname{Mixer}^{(l)}
=
\begin{cases}
    \operatorname{FullAttn}^{(l)}=A_{\mathrm{full}}^{(l)}, 
    & l \in \mathcal{I}_{\mathrm{full}}, \\
    \operatorname{LinearAttn}^{(l)}=A_{\mathrm{lin}}^{(l)}, 
    & l \notin \mathcal{I}_{\mathrm{full}} .
\end{cases}
\end{equation}

Therefore, a hybrid Transformer block can then be written as

\vspace{-0.6em}
\begin{equation}
\label{eq:hybrid_block}
\begin{aligned}
    \mathbf{H}^{(l)}
    &=
    \mathbf{X}^{(l)}
    +
    \operatorname{Mixer}^{(l)}
    \left(
    \operatorname{LN}(\mathbf{X}^{(l)})
    \right), \\
    \mathbf{X}^{(l+1)}
    &=
    \mathbf{H}^{(l)}
    +
    \operatorname{FFN}^{(l)}
    \left(
    \operatorname{LN}(\mathbf{H}^{(l)})
    \right).
\end{aligned}
\end{equation}

Here, the retained set \(\mathcal{I}_{\mathrm{full}}\) determines the core efficiency--effectiveness trade-off of the hybrid attention model. Keeping more full-attention layers preserves the retrieval and precise matching ability of the original Transformer, but also increases the computational and memory cost. Conversely, converting more layers to linear attention improves efficiency, but may degrade performance if critical layers are replaced. Therefore, under a fixed full-attention budget \(K = |\mathcal{I}_{\mathrm{full}}|\), the central question is not only how to train the linear-attention replacements, but also which layers should retain full attention, which is the focus of the next section.
\begin{figure*}[!t]
\centering
\includegraphics[width=\linewidth]{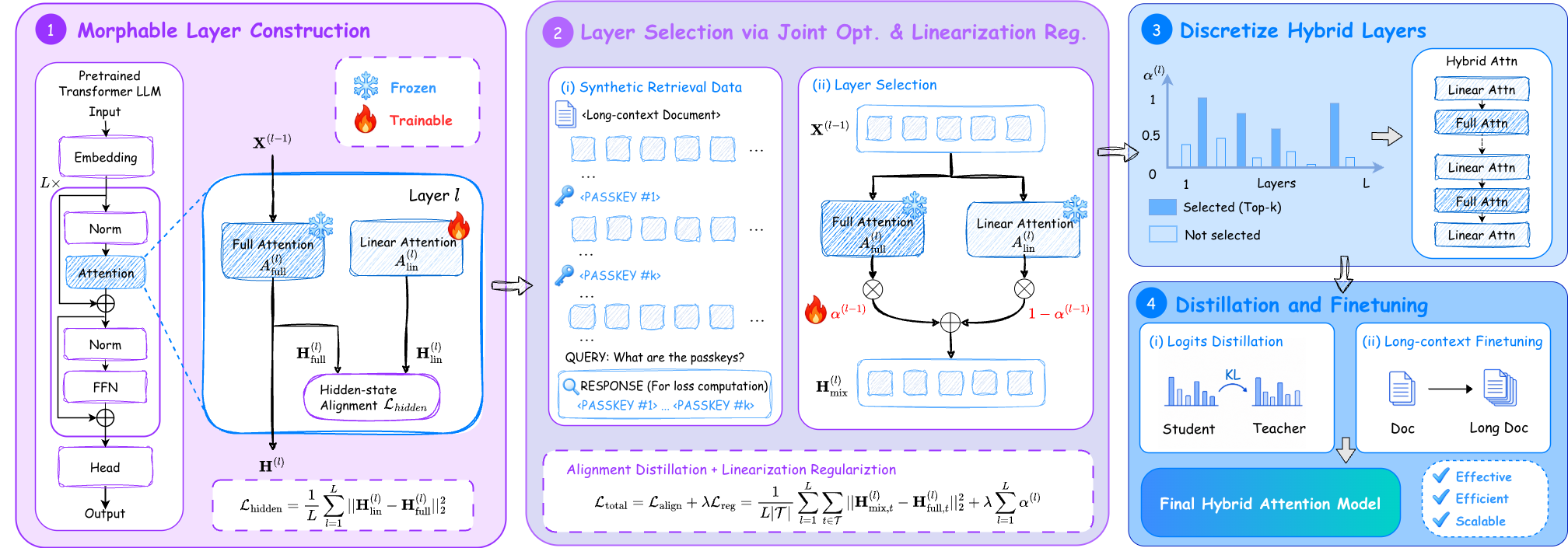}
\caption{\textbf{Overview of FlashMorph.} FlashMorph constructs morphable attention layers through hidden-state alignment, and performs layer selection by jointly optimizing one gate \(\alpha^{(l)}\) per layer with a linearization regularization on synthetic retrieval data. The learned gate values are discretized to retain the top-\(K\) full-attention layers and construct the hybrid attention model, followed by distillation and long-context finetuning. }
\label{fig:morph}
\end{figure*}

\section{Morphing into Hybrid Attention Models}

In this section, we introduce \textbf{FlashMorph} (\underline{\textbf{F}}ast \underline{\textbf{LA}}yer \underline{\textbf{S}}election for \underline{\textbf{H}}ybrid \underline{\textbf{MORPH}}ing), an effective, efficient, and scalable layer selection method for Transformer-to-hybrid conversion. As discussed in Sec.~\ref{sec:rethinking}, we formulate hybrid layer selection as a budget-constrained subset optimization problem, which motivates us to move beyond fixed allocation patterns and isolated layerwise scoring. As illustrated in Fig.~\ref{fig:morph}, FlashMorph first constructs morphable attention layers by distilling an all-linear model from the original Transformer through hidden-state alignment, thereby equipping each full-attention layer with a trained linear-attention replacement, as described in Sec.~\ref{sec:morphable}. It then performs layer selection by joint optimizing layerwise gate values on synthetic long-context retrieval data and discretizing the resulting gates into a hybrid attention model, as detailed in Sec.~\ref{sec:layer_selection}. Finally, in Sec.~\ref{sec:distillation} we apply standard logits distillation and long-context finetuning to further recover the quality of the hybrid model.

\vspace{-0.4em}
\subsection{Rethinking Layer Selection in Hybrid Attention Model}
\label{sec:rethinking}

Given a pretrained Transformer model with \(L\) attention layers, Transformer-to-hybrid conversion keeps only a subset of layers as full attention and converts the remaining layers to linear attention. Let \([L]=\{1,\dots,L\}\), and let \(\mathcal{I}_{\mathrm{full}}\subseteq [L]\) denote the retained full-attention layers. The corresponding hybrid model is denoted by \(\mathcal{M}(\mathcal{I}_{\mathrm{full}})\). Under a fixed full-attention budget \(K\), the ideal layer selection objective is 

\vspace{-0.6em}
\begin{equation} 
\label{eq:subset_selection} 
\mathcal{I}^{\star}_{\mathrm{full}} = \arg\max_{\substack{\mathcal{I}_{\mathrm{full}}\subseteq [L]\\|\mathcal{I}_{\mathrm{full}}|=K}} \operatorname{Score}\big(\mathcal{M}(\mathcal{I}_{\mathrm{full}})\big), 
\end{equation} 

where \(\operatorname{Score}(\cdot)\) is a higher-is-better evaluation metric used to represent the model's performance. This formulation shows that hybrid layer selection is inherently a subset optimization problem. Solving Eq.~\ref{eq:subset_selection} exactly requires evaluating \(\binom{L}{K}\) possible subsets, which is computationally infeasible for modern LLMs. 

Existing methods therefore rely on tractable approximations. Uniform interleaving~\citep{wang2024mamba,paliotta2025thinking,lan2025liger,goldstein2025radlads} imposes a fixed handcrafted pattern, while layerwise estimation~\citep{yang2026zebra,li2025distilling,chen2026hybrid} follows a cumbersome layer-by-layer procedure, where each layer is independently replaced, restored, or scored before the top-ranked layers are selected under the budget. This isolated evaluation is not only inefficient, but also implicitly treats layer importance as an isolated property while ignoring inter-layer dependencies and redundancy, causing the selected layers to be suboptimal when deployed as a joint hybrid configuration. We next introduce our proposed method FlashMorph for joint optimization-based hybrid layer selection.


\subsection{Morphable Layers Construction}
\label{sec:morphable}

We first construct a morphable model by equipping each full-attention layer with a converted linear-attention branch. To obtain these replacement branches, we follow prior Transformer-to-hybrid conversion pipelines~\citep{goldstein2025radlads,li2025distilling,chen2026hybrid}, where the pretrained full-attention model $\mathcal{M}_{\text{full}}$ serves as the teacher and the linear-attention branches are trained to imitate its layerwise representations.

During this stage, all parameters of the original full-attention model are frozen, and only the parameters of the linear-attention branches are updated. Let $\mathbf{H}_{\mathrm{full}}^{(l)}$ and $\mathbf{H}_{\mathrm{lin}}^{(l)}$ denote the output hidden states produced by the full-attention branch and the linear-attention branch at layer $l$, respectively. We optimize the linear-attention branches using a layerwise hidden-state alignment loss

\vspace{-0.6em}
\begin{equation}
\label{eq:linear_branch_training}
\mathcal{L}_{\mathrm{hidden}} = \frac{1}{L}
\sum_{l=1}^{L}\mathcal{L}_{\mathrm{hidden}}^{(l)} = 
\frac{1}{L}
\sum_{l=1}^{L}
\left\|
\mathbf{H}_{\mathrm{lin}}^{(l)} - \mathbf{H}_{\mathrm{full}}^{(l)}
\right\|_{2}^{2}.
\end{equation}

This stage yields a trained all-linear student $\mathcal{M}_{\text{all-linear}}$, which provides a linear-attention replacement for each full-attention layer in the original model. We then construct a morphable model by pairing these trained linear-attention replacements with the frozen full-attention layers of the pretrained model, enabling arbitrary full/linear layer configurations for subsequent layer selection stage.

\subsection{Layer Selection via Joint Optimization and Linearization Regularization}
\label{sec:layer_selection}

\textbf{Optimization-based layer selection.} Given the morphable model constructed above, layer selection aims to identify a subset of layers that should retain full attention under a fixed budget. This is a combinatorial subset-selection problem: for an $L$-layer model with a budget of $K$ retained full-attention layers, one must choose $K$ layers from $L$ candidates, resulting in $\binom{L}{K}$ possible hybrid configurations. Exhaustively evaluating these configurations is intractable. To avoid isolated layerwise scoring, FlashMorph relaxes the discrete subset-selection problem into a continuous optimization problem over layerwise gates, allowing all layers to be optimized jointly under a global hybrid configuration. Specifically, for each layer $l$, we introduce a scalar gate $\alpha^{(l)} \in [0,1]$ that interpolates between the full-attention branch and the linear-attention branch:

\vspace{-0.6em}
\begin{equation}
\label{eq:morph_gate}
\mathbf{H}^{(l)}_{\mathrm{mix}} = \alpha^{(l)} \mathbf{H}_{\mathrm{full}}^{(l)} + (1-\alpha^{(l)})\mathbf{H}_{\mathrm{lin}}^{(l)} .
\end{equation}

A larger $\alpha^{(l)}$ indicates stronger reliance on the full-attention branch, whereas a smaller $\alpha^{(l)}$ suggests that the layer can be more safely linearized. We initialize all gates as $\alpha^{(l)}=1$, corresponding to the original full-attention model. During layer selection stage, both the full-attention backbone and the trained linear-attention branches are frozen, and only the gate values $\boldsymbol{\alpha}=\{\alpha^{(l)}\}_{l=1}^{L}$ are optimized. This keeps the number of trainable parameters extremely small and prevents the selection stage from adapting model weights to compensate for a poor architectural choice.

To preserve the behavior of the original full-attention model, we optimize the gates by aligning the hidden states of the morphed model with those of the full-attention teacher. Let $\{\mathbf{H}_{\mathrm{full}}^{(l)}\}_{l=1}^{L}$ denote the hidden states of the original full-attention teacher, and let $\{\mathbf{H}_{\mathrm{mix}}^{(l)}\}_{l=1}^{L}$ denote the corresponding hidden states of the morphed model. We compute the alignment loss at the answer-token positions:

\vspace{-0.9em}
\begin{equation}
\label{eq:morph_alignment_loss}
\mathcal{L}_{\mathrm{align}} = \frac{1}{L|\mathcal{T}(x)|} \sum_{l=1}^{L}
\sum_{t\in \mathcal{T}(x)}
\left\|
\mathbf{H}_{\mathrm{mix},t}^{(l)} - \mathbf{H}_{\mathrm{full},t}^{(l)}
\right\|_{2}^{2},
\end{equation}

where $\mathcal{T}(x)$ denotes the set of answer-token positions. To encourage the model to rely on linear attention whenever for efficiency, we further introduce linearization regularization:

\vspace{-0.9em}
\begin{equation}
\label{eq:linearization_regularizer}
\mathcal{L}_{\mathrm{reg}} = \sum_{l=1}^{L}
\alpha^{(l)} .
\end{equation}

The final optimization objective for layer selection is therefore defined as

\vspace{-0.9em}
\begin{equation}
\label{eq:morph_selection_objective}
\mathcal{L}_{\mathrm{total}} = \mathcal{L}_{\mathrm{align}} + \lambda \mathcal{L}_{\mathrm{reg}},
\end{equation}

where $\lambda$ is a hyperparameter (set to 0.1 in our default implementation) controlling the strength of linearization regularization. The alignment term preserves the behavior of the full-attention teacher, while the regularization term penalizes reliance on full attention and pushes layers toward their linear-attention branches whenever this does not significantly change the teacher output. Since all gate values are optimized simultaneously, this objective enables FlashMorph to perform layer selection via joint optimization under a global hybrid configuration rather than scoring each layer in isolation. The resulting gates therefore reflect the relative necessity of retaining full attention while accounting for inter-layer dependency, redundancy, and complementarity.

\textbf{Synthetic retrieval data.} Generic language modeling objectives are often dominated by local dependencies and may provide weak supervision for identifying layers that are critical for long-context retrieval. We therefore perform layer selection on synthetic long-context retrieval examples. In each example, randomly generated passkeys are inserted at different depths of a long-context document, and the model is required to recover the corresponding passkeys at the end of the sequence. Similar synthetic retrieval signals have been used to identify long-context-critical attention components~\citep{xiao2025duoattention,lin2026lycheedecode,bick2026retrieval}. Implementation details are in Appendix.~\ref{app:imp}

In our setting, the synthetic retrieval data provides a targeted selection signal for measuring whether replacing full attention with linear attention disrupts long-range information access. Importantly, this data is used only to optimize the layer-wise gates rather than the model weights. As a result, the learned gates reflect how much each layer needs to retain full attention when all layers are optimized jointly, thereby capturing inter-layer dependency, redundancy, and complementarity that are overlooked by isolated layerwise estimation.

\textbf{Discretizing hybrid layers.} After optimization, the learned gate values indicate the relative necessity of preserving full attention in each layer. Given a target budget $K$ of full-attention layers, FlashMorph selects the layers as full attention with the largest gate values

\vspace{-0.9em}
\begin{equation}
\label{eq:morph_topk_selection}
\mathcal{I}_{\mathrm{full}}^{\mathrm{Hybrid}} = \operatorname{TopK} \left( \{\alpha_{l}\}_{l=1}^{L}, K \right).
\end{equation}

The selected layers $\mathcal{I}_{\mathrm{full}}^{\mathrm{Hybrid}}$ keep their original full-attention branches, while the remaining layers $\mathcal{I}_{\mathrm{lin}}^{\mathrm{Hybrid}}=[L]\setminus \mathcal{I}_{\mathrm{full}}^{\mathrm{Hybrid}}$ are instantiated with their trained linear-attention replacements. After this discretization step, the gate values are discarded, and the resulting architecture $\mathcal{M}(\mathcal{I}_{\mathrm{full}}^{\mathrm{Hybrid}})$ becomes a hybrid attention model with no additional overhead at inference time.

\begin{algorithm}[t]
\caption{FlashMorph Transformer-to-Hybrid Conversion}
\label{alg:morph}
\begin{algorithmic}[1]
\REQUIRE Full-attention model $\mathcal{M}_{\mathrm{full}}$ with $L$ layers; training dataset $\mathcal{D}$; synthetic retrieval dataset $\mathcal{D}_{\mathrm{syn}}$; layer selection optimization steps $S$; full-attention budget $K$; linearization regularization weight $\lambda$.

\STATE Distill an all-linear model $\mathcal{M}_{\mathrm{all\text{-}linear}}$ from $\mathcal{M}_{\mathrm{full}}$ on $\mathcal{D}$ by hidden-state alignment.
\STATE Construct a morphable model by equipping each full-attention branch $A_{\mathrm{full}}^{(l)}$ with its trained linear-attention replacement $A_{\mathrm{lin}}^{(l)}$.
\STATE Freeze both the full-attention model backbone and the linear-attention branches.
\STATE Initialize layerwise gates $\alpha^{(l)} \leftarrow 1$ for all $l=1,\dots,L$.

\FOR{$s=1,\dots,S$}
    \STATE Sample synthetic retrieval examples $x \sim \mathcal{D}_{\mathrm{syn}}$.
    \STATE Compute mixed hidden states for each layer:
    \[
    \mathbf{H}_{\mathrm{mix}}^{(l)}
    =
    \alpha^{(l)}\mathbf{H}_{\mathrm{full}}^{(l)} + \left(1-\alpha^{(l)}\right)\mathbf{H}_{\mathrm{lin}}^{(l)} .
    \]
    \STATE Compute the answer-token alignment loss $\mathcal{L}_{\mathrm{align}}$.
    \STATE Compute the linearization regularizer
    $
    \mathcal{L}_{\mathrm{reg}}=\sum_{l=1}^{L}\alpha^{(l)}.
    $
    \STATE Update only the gates $\boldsymbol{\alpha}$ by minimizing
    $
    \mathcal{L}_{\mathrm{total}} = \mathcal{L}_{\mathrm{align}} + \lambda \mathcal{L}_{\mathrm{reg}}.
    $
\ENDFOR

\STATE Select full-attention layers:
\[
\mathcal{I}_{\mathrm{full}}^{\mathrm{Hybrid}} = \operatorname{TopK}
\left(
\{\alpha^{(l)}\}_{l=1}^{L},
K
\right).
\]
\STATE Set linear-attention layers:
\[
\mathcal{I}_{\mathrm{lin}}^{\mathrm{Hybrid}}
=
[L]\setminus \mathcal{I}_{\mathrm{full}}^{\mathrm{Hybrid}} .
\]
\STATE Instantiate $\mathcal{M}_{\mathrm{Hybrid}}$ with full attention on $\mathcal{I}_{\mathrm{full}}^{\mathrm{Hybrid}}$ and linear attention on $\mathcal{I}_{\mathrm{lin}}^{\mathrm{Hybrid}}$; discard the gates $\boldsymbol{\alpha}$.
\STATE Apply logits distillation and long-context finetuning on $\mathcal{D}$.
\STATE \textbf{return} $\mathcal{M}_{\mathrm{Hybrid}}$.
\end{algorithmic}
\end{algorithm}

\vspace{-0.4em}
\subsection{Distillation and Long-context Finetuning} 
\label{sec:distillation}

After layer selection stage, we obtain a hybrid attention model $\mathcal{M}_{\mathrm{Hybrid}}=\mathcal{M}(\mathcal{I}_{\mathrm{full}}^{\mathrm{Hybrid}})$. Following prior Transformer-to-hybrid conversion pipelines~\citep{goldstein2025radlads,li2025distilling,chen2026hybrid}, we further apply logits distillation and long-context finetuning to recover the quality of the selected hybrid attention model.

\textbf{Logits distillation.} We distill the hybrid attention model from the original full-attention teacher. Let $p_{\mathrm{T}}(\cdot)$ and $p_{\mathrm{H}}(\cdot)$ denote the output logits generated by the full-attention teacher and the hybrid attention model, respectively. The distillation objective optimized by the Kullback-Leibler (KL) divergence $D_{\mathrm{KL}}$ is defined as

\vspace{-1em}
\begin{equation}
\label{eq:post_selection_kd}
\mathcal{L}_{\mathrm{KD}} = D_{\mathrm{KL}}
(p_{\mathrm{T}}(\mathbf{X}) \;\|\; p_{\mathrm{H}}(\mathbf{X})),
\end{equation}

\textbf{Long-context finetuning.}
We then finetune the hybrid model on long-context sequences with the standard language modeling objective

\vspace{-1em}
\begin{equation}
\label{eq:long_context_finetuning}
\mathcal{L}_{\mathrm{FT}} = -\sum\log p_{\mathrm{H}}(x_t \mid x_{<t}).
\end{equation}

This completes the FlashMorph Transformer-to-hybrid conversion pipeline. The full procedure is summarized in Algorithm~\ref{alg:morph}.

\section{Experiments}

\vspace{-0.3em}
\subsection{Experiment Setup}

\textbf{Baselines.} We compare FlashMorph with representative hybrid layer selection methods, including uniform interleaving, PostNAS~\citep{gu2025jet}, KL-LS~\citep{li2025distilling}, and HALO~\citep{chen2026hybrid}. Uniform interleaving retains full-attention layers according to a fixed periodic pattern, while PostNAS, KL-LS, and HALO represent data-driven selection strategies based on supernet training and searching, KL-guided scoring, or layerwise replacement evaluation. 

\textbf{Training.} We use Qwen3-0.6B and Qwen3-1.7B as the pretrained full-attention Transformer backbones. Throughout our experiments, we follow the HALO Transformer-to-hybrid conversion pipeline~\citep{chen2026hybrid}. Specifically, the pipeline first trains linear-attention replacement branches through hidden-state alignment, then applies layer selection to determine which layers retain full attention, and finally recovers the selected hybrid model through logits distillation and long-context finetuning. To isolate the effect of layer selection, we keep the model architecture, training data and pipeline unchanged across all methods, and replace only the layer selection strategy with FlashMorph or the corresponding baseline methods. All the experiments are conducted on $8\times$ GPUs with BFloat16 precision. Unless otherwise specified, all main comparisons are conducted under the 3:1 hybrid ratio. Our detailed model and training configurations are provided in Appendix~\ref{app:morph_config}.

\textbf{Evaluation.} We evaluate FlashMorph under two settings. For the Needle-in-a-Haystack (NIAH) task~\citep{hsieh2024ruler}, we follow the HypeNet setting~\citep{chen2026hybrid}, which adopts Lightning Attention~\citep{qin2024lightning} together with hybrid positional encoding (HyPE). The resulting hybrid model retains the key architectural components used in HypeNet, including attention logits scaling, QK normalization, GQA-to-MHA conversion, and output normalization and gating for linear-attention layers. Unlike the original HypeNet configuration, however, we do not apply gated attention~\citep{qiu2026gated} to the retained full-attention layers; instead, these layers remain standard full-attention blocks. For commonsense reasoning and recall-intensive tasks, we adopt the standard RoPE-based setting for all layers. Under this setting, we evaluate three linear-attention backbones: Lightning Attention~\citep{qin2024lightning}, Gated Linear Attention (GLA)~\citep{yang2023gated}, and Gated DeltaNet (GDN)~\citep{yang2024gated}. Our attention implementations are based on the \texttt{flash-linear-attention} library~\citep{yang2024fla}, and downstream evaluations are conducted using \texttt{lm-evaluation-harness}~\citep{eval-harness}.

\begin{table*}[t]
\centering
\caption{\textbf{NIAH Performance across 0.6B and 1.7B Backbones from 32K to 256K Context Lengths.} FlashMorph achieves strong retrieval performance with only 20M layer-selection tokens. $^{*}$ indicates selected layers taken from~\citep{chen2026hybrid}. The best results are highlighted in \textbf{bold}, and the second-best results are \underline{underlined}.} 
\small
\resizebox{\linewidth}{!}{
\begin{tabular}{lc|cccc|cccc|cccc}
    \toprule
    \multirow{2}{*}{\textbf{Model}} & \multirow{2}{*}{\textbf{\shortstack{LS. Tokens}}} & \multicolumn{4}{c}{\textbf{NIAH-Single-1}} & \multicolumn{4}{c}{\textbf{NIAH-Single-2}} & \multicolumn{4}{c}{\textbf{NIAH-Single-3}} \\
    \cmidrule(lr){3-14} 
    & & 32K & 64K & 128K & 256K & 32K & 64K & 128K & 256K & 32K & 64K & 128K & 256K  \\
    \midrule
    \multicolumn{14}{l}{\textit{0.6B backbone}} \\
    \rowcolor{gray!15} 
    Qwen3  & - & 100 & 100 & 0 & 0 & 100 & 99.0 & 0 & 0 & 99.8 & 82.8 & 0 & 0 \\
    \rowcolor{gray!15} 
    Qwen3+YaRN  & - & 30.2 & 29.2 & 20.2 & 31.0 & 0.6 & 0 & 0 & 0 & 0.8 & 0 & 0 & 0 \\
    \midrule
    Uniform & N/A & \underline{99.4} & 98.6 & 97.8 & \underline{98.0} & 68.4 & 49.2 & 41.2 & 12.8 & 57.6 & 36.0 & 28.8 & 12.2 \\
    KL-LS & 20B & 98.2 & 97.6 & 96.4 & 94.8 & 69.4 & 60.6 & \underline{50.2} & \textbf{24.0} & 32.2 & 17.2 & 8.6 & 3.6  \\
    HALO  & 234M & \textbf{99.6} & \textbf{99.8} & \textbf{99.6} & \textbf{99.2} & \underline{86.0} & \underline{80.6} & \textbf{62.4} & \underline{21.0} & \underline{68.6} & \underline{67.4} & \textbf{57.2} & \textbf{32.4} \\
    \rowcolor{cyan!20} 
    \textbf{FlashMorph}  & \textbf{20M} & 99.0 & \underline{99.0} & \underline{99.0} & \textbf{99.2} & \textbf{92.2} & \textbf{82.4} & 45.8 & 11.0 & \textbf{81.2} & \textbf{73.6} & \underline{45.6} & \underline{28.4} \\
    \midrule
    \midrule
    \multicolumn{14}{l}{\textit{1.7B backbone}} \\
    \rowcolor{gray!15} 
    Qwen3  & - & 100 & 100 & 0 & 0 & 100 & 98.8 & 0 & 0 & 99.8 & 96.4 & 0 & 0 \\
    \rowcolor{gray!15} 
    Qwen3+YaRN & - & 63.8 & 17.0 & 3.8 & 2.8 & 31.8 & 8.8 & 1.8 & 10.2 & 7.8 & 1.0 &  0.8 & 0.2  \\
    \midrule
    Uniform & N/A & \underline{99.8} & 99.6 & 99.6 & \textbf{100} & 71.8 & 86.8 & 28.4 & 19.2 & 58.6 & 59.4 & 16.4 & 27.8 \\
    PostNAS*  & 50B & 99.2 & 99.4 & \underline{99.8} & \underline{99.2} & 96.8 & 95.6 & 78.0 & 73.8  & 56.4 & 51.2 & 57.2 & \underline{57.6}  \\
    KL-LS  & 20B & 98.6 & 98.6 & 98.0 & 94.4 & 62.2 & 68.6 & 47.4 & 34.6 & 22.8 & 4.0 & 9.4 & 3.8 \\
    HALO  & 234M & \underline{99.8} & \textbf{100} & \textbf{100} & \textbf{100} & \textbf{99.6} & \underline{98.6} & \underline{95.0} & \textbf{95.2} & \underline{86.4} & \underline{90.8} & \underline{67.4} & 52.8 \\
    \rowcolor{cyan!20} 
    \textbf{FlashMorph}  & \textbf{20M} & \textbf{100} & \textbf{100} & \textbf{100} & \textbf{100} & \textbf{99.6} & \textbf{100} & \textbf{98.2} & \underline{88.2} & \textbf{96.6} & \textbf{95.4} & \textbf{94.4} & \textbf{73.2} \\
    \bottomrule
\end{tabular}
}
\addtolength{\tabcolsep}{2.5pt}    
\centering
\label{tab:niah}
\vspace{-1em}
\end{table*}

\vspace{-0.5em}
\subsection{Main Results}

\textbf{Needle-in-a-Haystack.} Table~\ref{tab:niah} reports the NIAH results on the 0.6B and 1.7B backbones across three retrieval variants and context lengths from 32K to 256K. The original Qwen3 models perform well at short context lengths but quickly collapse as the context is extended, while directly applying YaRN~\citep{peng2024yarn} fails to consistently restore long-context retrieval ability. Hybrid conversion substantially improves retrieval over extended contexts, but its effectiveness depends strongly on which layers are retained as full attention. On the 0.6B backbone, FlashMorph achieves near-perfect accuracy on NIAH-Single-1 and delivers strong performance on the more challenging NIAH-Single-2 and NIAH-Single-3 settings, particularly at short and medium context lengths. On the 1.7B backbone, the advantage becomes more pronounced: FlashMorph maintains perfect accuracy on NIAH-Single-1 across all context lengths, and substantially improves NIAH-Single-2 and NIAH-Single-3, where accurate retrieval over long contexts is more challenging. Notably, FlashMorph obtains these results using only 20M layer selection tokens, compared with substantially larger selection budgets required by prior methods. These results show that FlashMorph can identify full-attention layers effectively and efficiently while preserving strong long-context retrieval performance.

\begin{table*}[t]
\centering
\caption{\textbf{Zero-shot Performance on Commonsense Reasoning and Long-context Recall-intensive Tasks across Attention Backbones and Model Scales.} $^{*}$ indicates that the selected layers are derived from \citep{chen2026hybrid}. The best results are highlighted in \textbf{bold}, and the second-best results are \underline{underlined}.}
\label{tab:zero_shot}
\resizebox{\textwidth}{!}{
\begin{tabular}{llc|cccccc|cccc}
\toprule
\multirow{2}{*}{\textbf{Attn.}}
& \multirow{2}{*}{\textbf{Method}}
& \multirow{2}{*}{\textbf{LS. Tokens}}
& \textbf{ARC-e} & \textbf{ARC-c} & \textbf{PIQA} & \textbf{Hella.} & \textbf{Wino.}
& \multirow{2}{*}{\textbf{Avg.}}
& \textbf{SQuAD} & \textbf{FDA} & \textbf{SWDE} & \multirow{2}{*}{\textbf{Avg.}} \\
& & & acc & acc$_n$ & acc & acc$_n$ & acc & & acc & acc & acc & \\
\midrule

\multicolumn{13}{l}{\textit{0.6B backbone}} \\
\rowcolor{gray!15}
Full & Qwen3 & - & 60.6 & 34.1 & 67.6 & 47.3 & 55.7 & 53.1 & 44.1 & 82.1 & 80.5 & 68.9 \\

\midrule
\multirow{4}{*}{Lightning}
& Uniform & N/A & 62.3 & \underline{32.9} & 66.9 & 46.1 & \underline{55.6} & \underline{52.8} & 30.3 & 51.3 & 71.8 & 51.1 \\
& KL-LS & 20B  & 62.1 & \textbf{33.0} & \textbf{67.5} & \underline{46.2} & 54.9 & 52.7 & 29.3 & 58.5 & 70.5 & 52.8 \\
& HALO & 234M  & \textbf{63.5} & 32.4 & \underline{67.4} & \textbf{46.4} & \textbf{56.1} & \textbf{53.2} & \underline{34.7} & \underline{60.4} & \underline{72.6} & \underline{55.9} \\
\rowcolor{cyan!15}
& \textbf{FlashMorph}  & \textbf{20M}  & \underline{62.8} & 32.0 & 67.3 & \textbf{46.4} & 55.2 & 52.7 & \textbf{41.7} & \textbf{62.4} & \textbf{76.2} & \textbf{60.1} \\

\midrule
\multirow{4}{*}{GLA}
& Uniform & N/A & \underline{63.1} & \underline{33.1} & 67.3 & \underline{46.8} & \textbf{55.3} & \textbf{53.1} & 31.1 & 55.6 & 75.0 & 53.9 \\
& KL-LS  & 20B & 61.8 & \textbf{33.8} & 67.2 & \textbf{46.9} & 54.9 & 52.9 & 32.7 & 61.4 & 75.2 & 56.4 \\
& HALO  & 234M  & 62.9 & 32.6 & \underline{67.4} &\textbf{46.9} & \underline{55.1} & \underline{53.0} & \textbf{36.4} & \underline{68.5} & \underline{75.4} & \underline{60.1} \\
\rowcolor{cyan!15}
& \textbf{FlashMorph}  & \textbf{20M} & \textbf{63.9} & 32.9 & \textbf{67.5} & \textbf{46.9} & 54.1 & \textbf{53.1} & \underline{35.4} & \textbf{70.7} & \textbf{76.0} & \textbf{60.7} \\

\midrule
\multirow{4}{*}{GDN}
& Uniform & N/A & 59.6 & 32.3 & 67.1 & \textbf{47.6} & 55.3 & 52.4 & 30.3 & 55.5 & 72.2 & 52.7 \\
& KL-LS & 20B  & \underline{61.1} & \textbf{35.3} & \textbf{67.9} & 47.4 & \textbf{56.7} & \textbf{53.7} & \underline{33.5} & \textbf{72.8} & \underline{75.6} & \underline{60.6}  \\
& HALO  & 234M  & 60.1 & \underline{34.1} & 67.7 & \underline{47.5} & 55.3 & \underline{53.0} & 26.5 & 62.0 & 71.2 & 53.2 \\
\rowcolor{cyan!15}
& \textbf{FlashMorph}  & \textbf{20M} & \textbf{63.1} & 33.5 & \underline{67.8} & \underline{47.5} & \underline{56.4} & \textbf{53.7} & \textbf{38.4} & \underline{71.3} & \textbf{76.7} & \textbf{62.1} \\

\midrule
\midrule

\multicolumn{13}{l}{\textit{1.7B backbone}} \\
\rowcolor{gray!15}
Full & Qwen3 & - & 72.4 & 43.5 & 72.5 & 60.4 & 61.0 & 62.0 & 39.8 & 79.0 & 85.1 & 67.9 \\

\midrule
\multirow{5}{*}{Lightning}
& Uniform & N/A & \textbf{74.7} & \textbf{45.1} & 72.3 & 60.5 & \textbf{62.2} & \textbf{62.9} & 39.0 & 56.6 & 78.6 & 58.1 \\
& PostNAS* & 50B & \textbf{74.7} & \underline{43.0} & \underline{72.9} & \textbf{60.8} & 60.9 & \underline{62.5} & \textbf{51.3} & 64.7 & 80.9 & \underline{65.7} \\
& KL-LS  & 20B   & \underline{74.2} & 42.8 & 72.3 & 60.4 & 59.1 & 61.8 & 39.5 & 49.4 & 74.8 & 54.5 \\
& HALO & 234M    & 73.5 & 42.8 & 72.6 & \underline{60.6} & \underline{61.5} & 62.2 & \textbf{51.3} & \underline{70.5} & \textbf{82.5} & \textbf{68.1} \\
\rowcolor{cyan!15}
& \textbf{FlashMorph}  & \textbf{20M} & 73.2 & 42.3 & \textbf{73.2} & \underline{60.6} & 61.0 & 62.1 & \underline{51.1} & \textbf{71.6} & \underline{81.6} & \textbf{68.1}  \\

\midrule
\multirow{5}{*}{GLA}
& Uniform & N/A & 74.4 & \underline{45.5} & 72.7 & 60.8 & \textbf{63.0} & \textbf{63.3} & 43.6 & 57.3 & 81.2 & 60.7 \\
& PostNAS* & 50B & 73.4 & 43.1 & \underline{73.2} & \textbf{61.5} & 60.1 & 62.3 & \textbf{54.2} & \underline{69.4} & \underline{81.9} & \textbf{68.5} \\
& KL-LS  & 20B   & \textbf{74.6} & \textbf{46.3} & 72.4 & 61.0 & 59.8 & 62.8 & 44.1 & 53.5 & 78.6 & 58.7 \\
& HALO   & 234M  & 74.4 & 44.5 & 72.6 & 60.9 & \underline{62.8} & 63.0 & 43.7 & 41.4 & 74.2 & 53.1 \\
\rowcolor{cyan!15}
& \textbf{FlashMorph}  & \textbf{20M}  & \underline{74.5} & 44.7 & \textbf{73.3} & \underline{61.2} & 62.3 & \underline{63.2} & \underline{47.3} & \textbf{73.7} & \textbf{82.3} & \underline{67.7} \\

\midrule
\multirow{5}{*}{GDN}
& Uniform & N/A  & 74.3 & 44.5 & 72.3 & 61.5 & 62.4 & 63.0 & 48.2 & 59.2 & 78.9 & 62.1 \\
& PostNAS* & 50B & \underline{75.1} & 45.1 & 73.1 & \textbf{61.7} & 62.3 & 63.4 & 52.8 & 67.9 & \textbf{82.4} & 67.7 \\
& KL-LS  & 20B  & 74.3 & \textbf{47.6} & 72.9 & \underline{61.6} & 62.3 & \underline{63.7} & 54.1 & \underline{73.5} & 81.2 & \underline{69.6} \\
& HALO  & 234M  & \textbf{75.8} & \underline{45.7} & \underline{73.2} & 61.1 & \underline{63.1} & \textbf{63.8} & \textbf{54.5} & 68.2 & 80.9 & 67.9 \\
\rowcolor{cyan!15}
& \textbf{FlashMorph}  & \textbf{20M}  & 74.4 & 44.1 & \textbf{73.3} & 61.5 & \textbf{63.9} & 63.4 & \underline{54.3} & \textbf{74.1} & \underline{82.2} & \textbf{70.2} \\

\bottomrule
\end{tabular}
}
\end{table*}

\textbf{Commonsense Reasoning and Recall-intensive Tasks.} We evaluate zero-shot commonsense reasoning tasks, including ARC-Easy (ARC-e) and ARC-Challenge (ARC-c)~\citep{clark2018think}, PIQA~\citep{bisk2020piqa}, HellaSwag (Hella.)~\citep{zellers2019hellaswag}, and WinoGrande (Wino.)~\citep{sakaguchi2021winogrande}; as well as real-world recall-intensive tasks, including SQuAD~\citep{rajpurkar2018know}, FDA~\citep{arora2023language}, and SWDE~\citep{lockard2019openceres}. As shown in Table~\ref{tab:zero_shot}, FlashMorph preserves strong zero-shot commonsense performance across attention backbones and model scales. On the 0.6B backbone, it achieves competitive commonsense averages under Lightning Attention and GLA, and matches the best average performance under GDN. On the 1.7B backbone, FlashMorph remains close to the strongest baselines across all three attention backbones, suggesting that the selected hybrid architectures largely retain the general reasoning ability of the original Transformer. The advantage of FlashMorph is more pronounced on recall-intensive tasks. On the 0.6B backbone, FlashMorph achieves the highest recall average across all three attention backbones. On the 1.7B backbone, it obtains the best or tied-best recall average with Lightning Attention and GDN, and remains highly competitive with GLA, where it reaches 67.7 compared with 68.5 from PostNAS while using orders of magnitude fewer layer-selection tokens. Overall, these results demonstrate that joint optimization-based layer selection provides an effective and efficient way to preserve recall capabilities while maintaining general performance.

\subsection{Efficiency Results}

\begin{figure}[!t]
\centering
\includegraphics[width=0.96\linewidth]{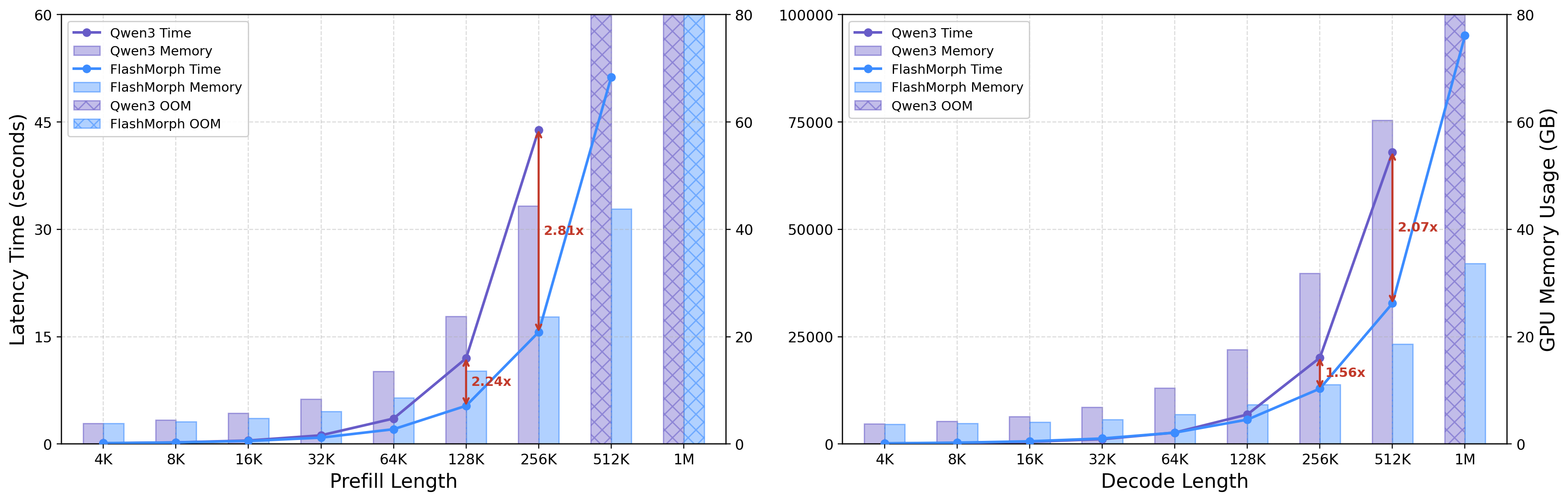}
\vspace{-1em}
\caption{\textbf{Prefilling and Decoding Efficiency Comparison.} FlashMorph achieves substantially better long-context efficiency than Qwen3. For prefill, FlashMorph becomes increasingly faster as context length grows, reaching 2.24$\times$ speedup at 128K and 2.81$\times$ at 256K. For decode, FlashMorph also shows clear advantages at long lengths, with 1.56$\times$ speedup at 256K and 2.07$\times$ at 512K, while using much less GPU memory, demonstrating the efficiency of hybrid attention model. Hatched bars indicate out of memory (OOM).}
\label{fig:efficiency}
\vspace{-1em}
\end{figure}

We evaluate the inference efficiency of FlashMorph (linear:full=3:1 hybrid attention) and Qwen3 (purely full attention) based on 1.7B backbone on single GPU. As shown in Fig.~\ref{fig:efficiency}, we report both latency time and peak GPU memory usage under increasing sequence lengths for prefilling and decoding, with a fixed batch size of 1.

\begin{figure}[!t]
\centering
\includegraphics[width=1.0\linewidth]{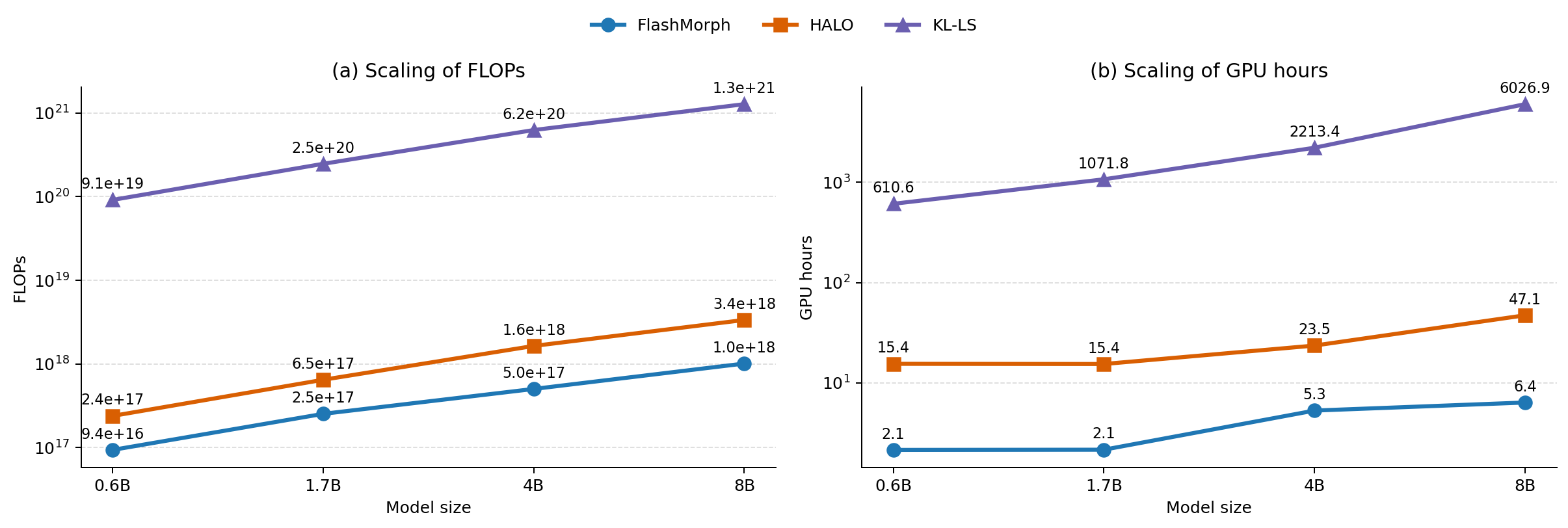}
\vspace{-2em}
\caption{\textbf{Efficiency Scaling of Layer Selection Methods across Model Sizes}. FlashMorph consistently requires substantially fewer FLOPs and GPU hours than HALO and KL-LS, with the efficiency gap becoming more pronounced as model size increases.}
\label{fig:layer_selection_scale}
\vspace{-1.3em}
\end{figure}

\textbf{Prefilling.} For prefilling, we vary the input length from 4K to 1M tokens. FlashMorph achieves comparable latency to Qwen3 at short sequence lengths, while its advantage becomes increasingly pronounced as the context length grows. Specifically, FlashMorph achieves a $2.24\times$ speedup at 128K tokens and a $2.81\times$ speedup at 256K tokens. It also consistently uses less GPU memory, enabling it to scale to 512K-token prefilling on a single GPU, whereas Qwen3-1.7B encounters out-of-memory at this length.

\textbf{Decoding.} For decoding, we fix the prefilling length to 1K tokens and vary the decoding length from 4K to 1M tokens. FlashMorph shows a flatter growth trend in both latency and GPU memory usage as the decoding length increases. It achieves a $1.56\times$ speedup at 256K tokens and a $2.07\times$ speedup at 512K tokens. Moreover, FlashMorph remains executable at 1M decoding length, while Qwen3-1.7B runs out of memory, demonstrating the improved long-context scalability of the hybrid attention model architecture.

\begin{wraptable}{r}{0.43\linewidth}
\vspace{-1.3em}
\caption{\textbf{Efficiency Comparison of Layer Selection Methods}. FlashMorph performs joint optimization-based layer selection with only 20M tokens and 2.1 GPU hours, substantially reducing the cost compared with prior search-based and layerwise methods. *The result is calculated from~\citep{gu2025jet}.}
\label{tab:layer_selection_cost}
\small
\setlength{\tabcolsep}{3.5pt}
\resizebox{\linewidth}{!}{
\begin{tabular}{lccc}
\toprule
\textbf{LS Method} & \textbf{Tokens $\downarrow$} & \textbf{FLOPs  $\downarrow$} & \textbf{GPU hours $\downarrow$} \\
\midrule
PostNAS* & 50B & 8.0e20 & 2561.3 \\
KL-LS & 20B & 2.5e20 & 1071.8 \\
HALO & 234M & 6.5e17 & 15.4 \\
\textbf{FlashMorph } & \textbf{20M} & \textbf{2.5e17} & \textbf{2.1} \\
\bottomrule
\end{tabular}
}
\end{wraptable}

\begin{figure}[!t]
\centering

\begin{subfigure}[t]{0.59\linewidth}
    \centering
    \includegraphics[width=\linewidth]{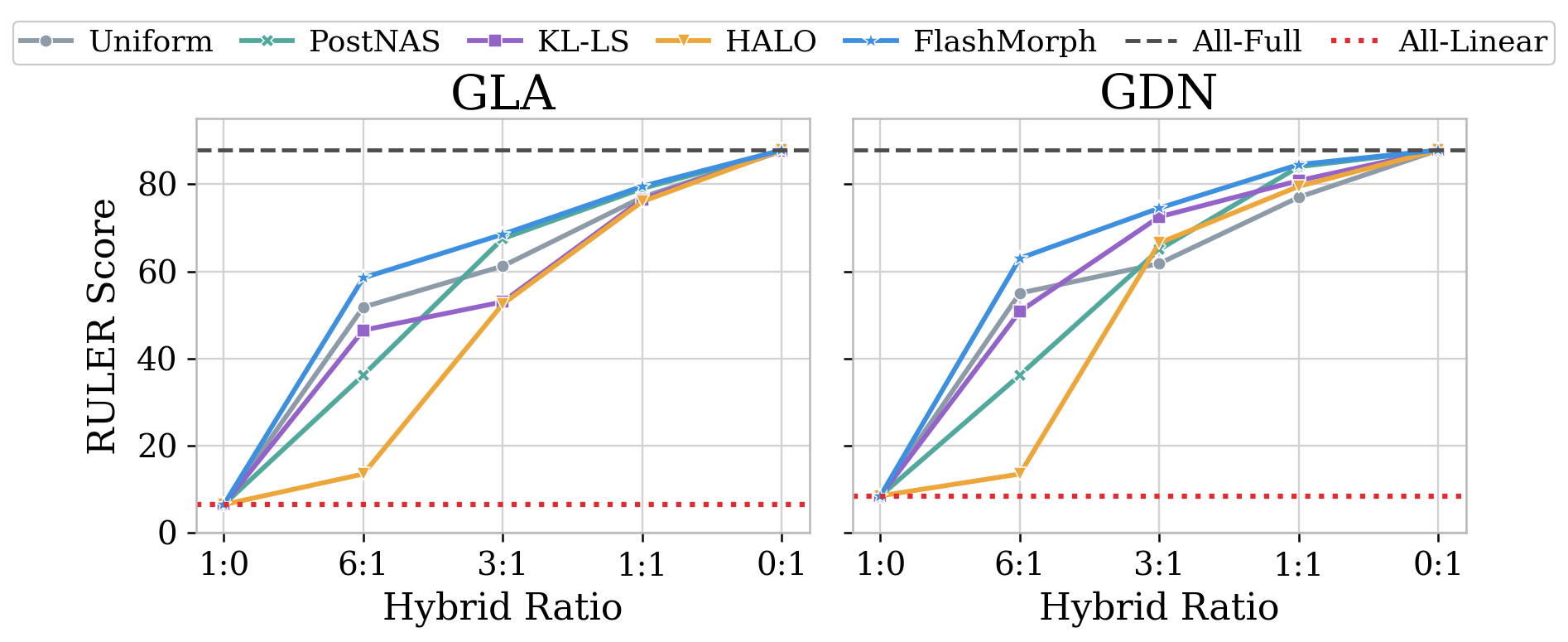}
    \caption{RULER performance under different hybrid ratios.}
    \label{fig:hybrid_ratio}
\end{subfigure}
\hfill
\begin{subfigure}[t]{0.37\linewidth}
    \centering
    \includegraphics[width=\linewidth]{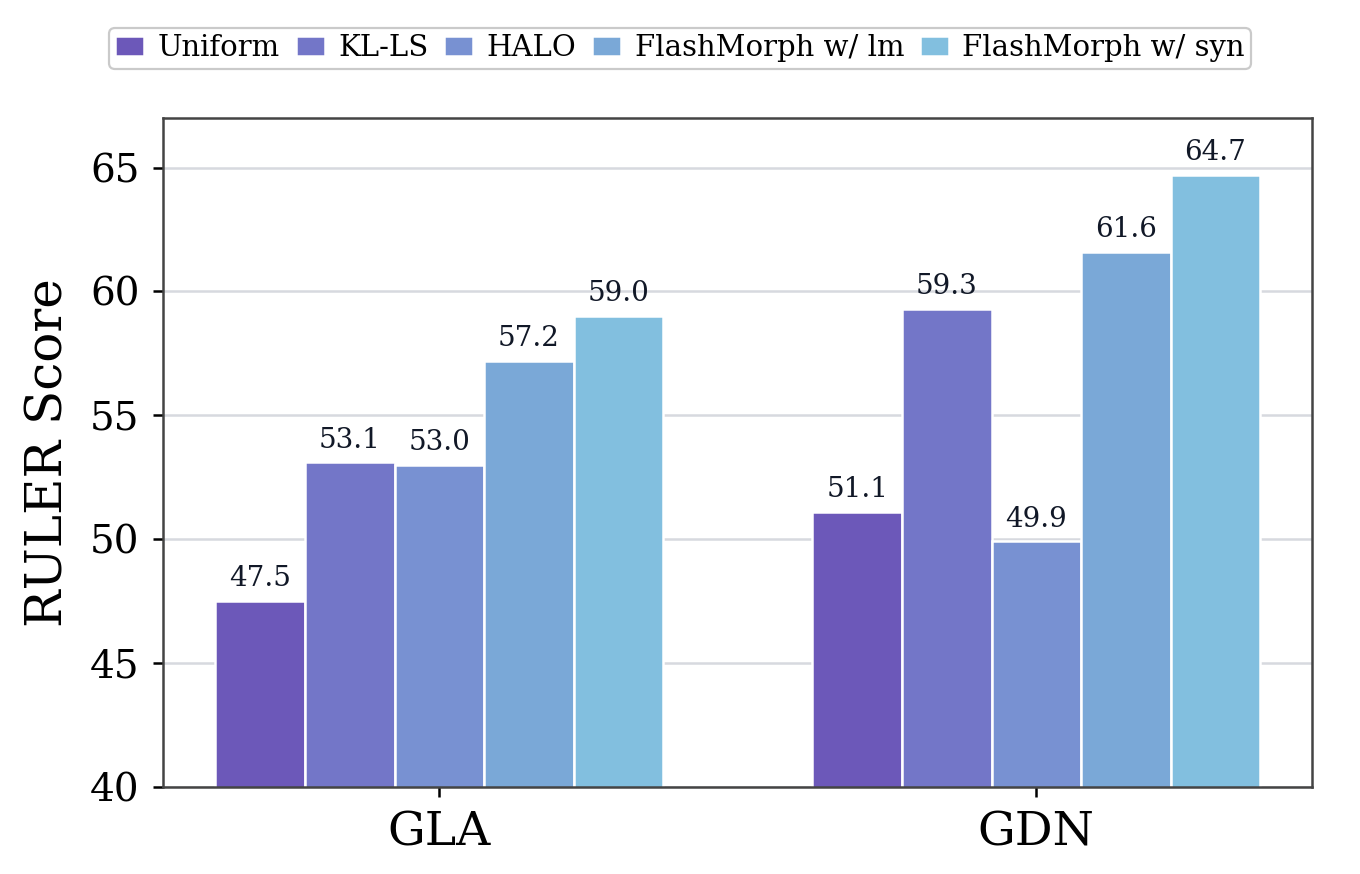}
    \caption{Effect of layer selection supervision.}
    \label{fig:ablation}
\end{subfigure}

\vspace{-0.5em}
\caption{\textbf{Analysis of FlashMorph under Different Hybrid Configurations and Supervision Signals.}
(a) We compare FlashMorph with prior layer selection methods on GLA and GDN backbones under different hybrid ratios. FlashMorph consistently achieves strong performance across hybrid ratios.
(b) FlashMorph with language-modeling supervision already outperforms prior layer selection methods, while synthetic passkey-based supervision further improves performance on both GLA and GDN backbones.}
\label{fig:analysis}
\vspace{-1.2em}
\end{figure}

\textbf{Layer selection.} We compare the layer selection cost of FlashMorph with existing layer selection methods in terms of required tokens, FLOPs, and GPU hours. Table~\ref{tab:layer_selection_cost} reports the results based on Qwen3-1.7B. FlashMorph uses only 20M tokens for hybrid layer selection, requiring $2.5\times10^{17}$ FLOPs and 2.1 GPU hours. This is substantially lower than PostNAS, KL-LS, and HALO, which require 50B, 20B, and 234M tokens, respectively. In terms of GPU hours, FlashMorph reduces the selection cost by $7.3\times$ compared with HALO, $510.4\times$ compared with KL-LS, and $1219.7\times$ compared with PostNAS. Figure~\ref{fig:layer_selection_scale} further compares the scaling behavior of different layer selection methods across model sizes. KL-LS scales poorly because it repeatedly restores and distills single-layer variants, leading to rapidly increasing FLOPs and GPU hours as the model size grows. HALO is more efficient but still requires layerwise evaluation. In contrast, FlashMorph optimizes layer-wise gates under a global hybrid configuration, making the selection stage lightweight and scalable. As a result, FlashMorph consistently achieves the lowest FLOPs and GPU hours across all model sizes, demonstrating the efficiency of joint optimization-based layer selection.

\vspace{-0.5em}
\subsection{Analysis}

\textbf{Robustness across Hybrid Ratios.} We further evaluate whether FlashMorph remains effective under different hybrid ratios based on the Qwen3-1.7B backbone, including linear:full hybrid settings of $6{:}1$, $3{:}1$, and $1{:}1$. For each ratio, we keep the post-selection distillation and finetuning pipeline unchanged and vary only the layer selection strategy, thereby isolating the effect of full-attention layer allocation under different full-attention budgets. As shown in Fig.~\ref{fig:hybrid_ratio}, FlashMorph consistently achieves strong RULER~\citep{hsieh2024ruler} performance across both GLA and GDN backbones. When the full-attention budget is limited, the advantage of FlashMorph is particularly pronounced: at the $6{:}1$ ratio, FlashMorph substantially outperforms prior selection methods on both backbones, indicating that its selected full-attention layers are more effective under sparse full-attention allocation. As the full-attention budget increases from $6{:}1$ to $1{:}1$, all methods improve and gradually approach the all-full upper bound, while FlashMorph remains the best or among the strongest methods across ratios. These results show that FlashMorph is not tied to a single hybrid budget, but provides robust layer selection under different efficiency-performance trade-offs in long-context settings.

\textbf{Comparison of Layer Selection Supervision.} We investigate how the supervision signal used during layer selection affects the resulting hybrid configuration. We compare two variants of FlashMorph based on Qwen3-0.6B backbone: one optimizes the layerwise gates with standard language-modeling supervision on generic text data, and the other uses our synthetic retrieval data with passkey answer-token alignment. As shown in Figure~\ref{fig:ablation}, FlashMorph with language modeling supervision (FlashMorph w/ lm) already outperforms prior layer selection methods, including KL-LS and HALO on both GLA and GDN backbones. This indicates that the proposed joint optimization can identify stronger hybrid configurations even without specialized retrieval-oriented data. Nevertheless, using synthetic retrieval data supervision (FlashMorph w/ syn) further improves the RULER score from 57.2 to 59.0 on GLA and from 61.6 to 64.7 on GDN. These gains suggest that retrival-oriented supervision more directly stresses long-context recall and better identifies retrieval-critical full-attention layers compared with standard language modeling supervision. Overall, the results show that FlashMorph benefits from both joint optimization and retrieval-oriented supervision, with the synthetic passkey objective providing the strongest layer selection signal in long-context settings.

\section{Related Work}

\subsection{Linear RNN and Hybrid Attention} 

To address the quadratic computational cost and massive memory usage of full attention, a variety of linear RNN architectures have been proposed to enable more efficient decoding and scalable long-context processing~\citep{katharopoulos2020transformers,peng2023rwkv,gu2023mamba,dao2024transformers,qin2023hierarchically,qin2024lightning,chou2024metala,sun2025linear,zhang2024gated,yang2024parallelizing,yang2024gated,du2025mom,lahoti2026mamba,hu2025comba}. However, entirely replacing full attention with linear RNN-style sequence mixers can introduce a fixed-state memory bottleneck, limiting the model’s capacity for recall-intensive operations such as associative retrieval and long-range information access~\citep{arora2024simple,wen2025rnns}.

Rather than replacing full attention entirely, hybrid attention models combine full attention with linear RNN  mixers~\citep{lieber2024jamba,glorioso2024zamba,de2024griffin,ren2025samba,chen2025minimax,zuo2025falcon,blakeman2025nemotron,wang2025systematic,du2025native,qwen_qwen3_coder_next_tech_report,team2025kimi_linear,qwen3.5,merrill2026olmo}. By retaining a small subset of full-attention layers, these models preserve global information access, while using linear RNN layers to improve inference efficiency. However, most existing hybrid architectures are designed and pretrained from scratch, typically relying on fixed allocation patterns such as uniform interleaving of full-attention and linear-attention layers. Although effective in the pretraining setting, such handcrafted patterns do not directly address the layer selection problem that arises when converting an existing pretrained Transformer into a hybrid attention model: determining which layers should retain full attention and which can be safely linearized.

\subsection{Transformer-to-hybrid Conversion}

Converting pretrained Transformers into efficient hybrid attention models has emerged as an appealing alternative to training from scratch. Existing conversion pipelines typically replace full-attention layers with linear-attention through weight transfer, followed by distillation and continued finetuning to recover model quality~\citep{kasai2021finetuning,mercat2024linearizing,chen2024dijiang,zhang2024hedgehog,wang2024mamba,bick2024transformers,lan2025liger,bick2025llamba,paliotta2025thinking,goldstein2025radlads,li2025distilling,chen2026hybrid}. In the Transformer-to-hybrid conversion setting, only a limited number of layers can retain full attention, while the remaining layers are linearized. This makes layer selection a central challenge: under a fixed full-attention budget, the conversion process must determine which layers should preserve full attention and which layers can be safely replaced by linear attention.

Existing hybrid layer selection methods typically rely on uniform interleaving or layerwise importance estimation. Uniform interleaving is simple and architecture-agnostic, but it ignores the non-uniform functional roles of pretrained Transformer layers. Search-based methods such as PostNAS~\citep{gu2025jet} move beyond purely layerwise scoring, but require training a supernet, which introduces substantial search overhead. Layerwise methods estimate marginal utility of each layer by perturbing, replacing, restoring, and scoring one layer at a time, as in SMART~\citep{yang2026zebra}, KL-LS~\citep{li2025distilling}, and HALO~\citep{chen2026hybrid}. Although practical, these layerwise procedures are cumbersome and treat layer importance in isolation, thereby overlooking interdependent layer effect. In contrast, FlashMorph formulates hybrid layer selection as a joint optimization problem. By learning layer-wise gates within a morphable model via linearization regularization, FlashMorph accounts for inter-layer dependencies, redundancy, and complementarity under a global hybrid configuration, while simultaneously avoiding the repeated layer-by-layer replacement and evaluation required by prior methods, which makes the layer selection stage substantially more efficient.

\section{Conclusion}

In this paper, we presented \textbf{FlashMorph}, an effective, efficient, and scalable layer selection method for converting pretrained Transformers into hybrid attention models. Rather than relying on fixed placement rules or isolated layerwise scoring, FlashMorph formulates hybrid layer selection as a budget-constrained joint optimization problem that accounts for inter-layer dependencies, redundancy, and complementarity, and constructs morphable attention layers to optimize lightweight layerwise gates under a global hybrid configuration. Extensive experiments across Qwen3-series backbones and multiple linear-attention variants show that FlashMorph preserves strong long-context retrieval and recall-intensive performance, maintains competitive commonsense reasoning ability, and substantially reduces layer selection overhead, demonstrating the effectiveness, efficiency, and scalability of joint optimization-based Transformer-to-hybrid conversion.

\clearpage

\bibliographystyle{plainnat}
\bibliography{reference}

\begin{thebibliography}{72}
\providecommand{\natexlab}[1]{#1}
\providecommand{\url}[1]{\texttt{#1}}
\expandafter\ifx\csname urlstyle\endcsname\relax
  \providecommand{\doi}[1]{doi: #1}\else
  \providecommand{\doi}{doi: \begingroup \urlstyle{rm}\Url}\fi

\bibitem[Arora et~al.(2023)Arora, Yang, Eyuboglu, Narayan, Hojel, Trummer, and R{\'e}]{arora2023language}
Simran Arora, Brandon Yang, Sabri Eyuboglu, Avanika Narayan, Andrew Hojel, Immanuel Trummer, and Christopher R{\'e}.
\newblock Language models enable simple systems for generating structured views of heterogeneous data lakes.
\newblock \emph{arXiv preprint arXiv:2304.09433}, 2023.

\bibitem[Arora et~al.(2024)Arora, Eyuboglu, Zhang, Timalsina, Alberti, Zinsley, Zou, Rudra, and R{\'e}]{arora2024simple}
Simran Arora, Sabri Eyuboglu, Michael Zhang, Aman Timalsina, Silas Alberti, Dylan Zinsley, James Zou, Atri Rudra, and Christopher R{\'e}.
\newblock Simple linear attention language models balance the recall-throughput tradeoff.
\newblock \emph{arXiv preprint arXiv:2402.18668}, 2024.

\bibitem[Bick et~al.(2024)Bick, Li, Xing, Kolter, and Gu]{bick2024transformers}
Aviv Bick, Kevin~Y Li, Eric~P Xing, J~Zico Kolter, and Albert Gu.
\newblock Transformers to ssms: Distilling quadratic knowledge to subquadratic models.
\newblock \emph{Advances in neural information processing systems}, 37:\penalty0 31788--31812, 2024.

\bibitem[Bick et~al.(2025)Bick, Katsch, Sohoni, Desai, and Gu]{bick2025llamba}
Aviv Bick, Tobias Katsch, Nimit Sohoni, Arjun Desai, and Albert Gu.
\newblock Llamba: Scaling distilled recurrent models for efficient language processing.
\newblock \emph{arXiv preprint arXiv:2502.14458}, 2025.

\bibitem[Bick et~al.(2026)Bick, Xing, and Gu]{bick2026retrieval}
Aviv Bick, Eric~P Xing, and Albert Gu.
\newblock Retrieval-aware distillation for transformer-ssm hybrids.
\newblock \emph{arXiv preprint arXiv:2602.11374}, 2026.

\bibitem[Bisk et~al.(2020)Bisk, Zellers, Gao, Choi, et~al.]{bisk2020piqa}
Yonatan Bisk, Rowan Zellers, Jianfeng Gao, Yejin Choi, et~al.
\newblock Piqa: Reasoning about physical commonsense in natural language.
\newblock In \emph{Proceedings of the AAAI conference on artificial intelligence}, volume~34, pages 7432--7439, 2020.

\bibitem[Blakeman et~al.(2025)Blakeman, Basant, Khattar, Renduchintala, Bercovich, Ficek, Bjorlin, Taghibakhshi, Deshmukh, Mahabaleshwarkar, et~al.]{blakeman2025nemotron}
Aaron Blakeman, Aarti Basant, Abhinav Khattar, Adithya Renduchintala, Akhiad Bercovich, Aleksander Ficek, Alexis Bjorlin, Ali Taghibakhshi, Amala~Sanjay Deshmukh, Ameya~Sunil Mahabaleshwarkar, et~al.
\newblock Nemotron-h: A family of accurate and efficient hybrid mamba-transformer models.
\newblock \emph{arXiv preprint arXiv:2504.03624}, 2025.

\bibitem[Cai et~al.(2024)Cai, Cao, Chen, Chen, Chen, Chen, Chen, Chen, Chen, Chu, et~al.]{cai2024internlm2}
Zheng Cai, Maosong Cao, Haojiong Chen, Kai Chen, Keyu Chen, Xin Chen, Xun Chen, Zehui Chen, Zhi Chen, Pei Chu, et~al.
\newblock Internlm2 technical report.
\newblock \emph{arXiv preprint arXiv:2403.17297}, 2024.

\bibitem[Chen et~al.(2025)Chen, Li, Gong, Jiang, Fei, Yang, Shan, Yu, Wang, Zhu, et~al.]{chen2025minimax}
Aili Chen, Aonian Li, Bangwei Gong, Binyang Jiang, Bo~Fei, Bo~Yang, Boji Shan, Changqing Yu, Chao Wang, Cheng Zhu, et~al.
\newblock Minimax-m1: Scaling test-time compute efficiently with lightning attention.
\newblock \emph{arXiv preprint arXiv:2506.13585}, 2025.

\bibitem[Chen et~al.(2024)Chen, Liu, Wang, Tian, and Wang]{chen2024dijiang}
Hanting Chen, Zhicheng Liu, Xutao Wang, Yuchuan Tian, and Yunhe Wang.
\newblock Dijiang: Efficient large language models through compact kernelization.
\newblock \emph{arXiv preprint arXiv:2403.19928}, 2024.

\bibitem[Chen et~al.(2026)Chen, Thai, Zhou, Zhang, Shen, Wang, Xiao, Han, and Liu]{chen2026hybrid}
Yingfa Chen, Zhen~Leng Thai, Zihan Zhou, Zhu Zhang, Xingyu Shen, Shuo Wang, Chaojun Xiao, Xu~Han, and Zhiyuan Liu.
\newblock Hybrid linear attention done right: Efficient distillation and effective architectures for extremely long contexts.
\newblock \emph{arXiv preprint arXiv:2601.22156}, 2026.

\bibitem[Chou et~al.(2024)Chou, Yao, Wang, Pan, Zhu, Zhong, Qiao, Wu, Xu, and Li]{chou2024metala}
Yuhong Chou, Man Yao, Kexin Wang, Yuqi Pan, Ruijie Zhu, Yiran Zhong, Yu~Qiao, Jibin Wu, Bo~Xu, and Guoqi Li.
\newblock Metala: Unified optimal linear approximation to softmax attention map.
\newblock \emph{Advances in Neural Information Processing Systems}, 37:\penalty0 71034--71067, 2024.

\bibitem[Clark et~al.(2018)Clark, Cowhey, Etzioni, Khot, Sabharwal, Schoenick, and Tafjord]{clark2018think}
Peter Clark, Isaac Cowhey, Oren Etzioni, Tushar Khot, Ashish Sabharwal, Carissa Schoenick, and Oyvind Tafjord.
\newblock Think you have solved question answering? try arc, the ai2 reasoning challenge.
\newblock \emph{arXiv preprint arXiv:1803.05457}, 2018.

\bibitem[Dao and Gu(2024)]{dao2024transformers}
Tri Dao and Albert Gu.
\newblock Transformers are ssms: Generalized models and efficient algorithms through structured state space duality.
\newblock \emph{arXiv preprint arXiv:2405.21060}, 2024.

\bibitem[De et~al.(2024)De, Smith, Fernando, Botev, Cristian-Muraru, Gu, Haroun, Berrada, Chen, Srinivasan, et~al.]{de2024griffin}
Soham De, Samuel~L Smith, Anushan Fernando, Aleksandar Botev, George Cristian-Muraru, Albert Gu, Ruba Haroun, Leonard Berrada, Yutian Chen, Srivatsan Srinivasan, et~al.
\newblock Griffin: Mixing gated linear recurrences with local attention for efficient language models.
\newblock \emph{arXiv preprint arXiv:2402.19427}, 2024.

\bibitem[Du et~al.(2025{\natexlab{a}})Du, Hu, Zhang, Sun, and Cheng]{du2025native}
Jusen Du, Jiaxi Hu, Tao Zhang, Weigao Sun, and Yu~Cheng.
\newblock Native hybrid attention for efficient sequence modeling.
\newblock \emph{arXiv preprint arXiv:2510.07019}, 2025{\natexlab{a}}.

\bibitem[Du et~al.(2025{\natexlab{b}})Du, Sun, Lan, Hu, and Cheng]{du2025mom}
Jusen Du, Weigao Sun, Disen Lan, Jiaxi Hu, and Yu~Cheng.
\newblock Mom: Linear sequence modeling with mixture-of-memories.
\newblock \emph{arXiv preprint arXiv:2502.13685}, 2025{\natexlab{b}}.

\bibitem[Gao et~al.(2024)Gao, Tow, Abbasi, Biderman, Black, DiPofi, Foster, Golding, Hsu, Le~Noac'h, Li, McDonell, Muennighoff, Ociepa, Phang, Reynolds, Schoelkopf, Skowron, Sutawika, Tang, Thite, Wang, Wang, and Zou]{eval-harness}
Leo Gao, Jonathan Tow, Baber Abbasi, Stella Biderman, Sid Black, Anthony DiPofi, Charles Foster, Laurence Golding, Jeffrey Hsu, Alain Le~Noac'h, Haonan Li, Kyle McDonell, Niklas Muennighoff, Chris Ociepa, Jason Phang, Laria Reynolds, Hailey Schoelkopf, Aviya Skowron, Lintang Sutawika, Eric Tang, Anish Thite, Ben Wang, Kevin Wang, and Andy Zou.
\newblock The language model evaluation harness, 07 2024.
\newblock URL \url{https://zenodo.org/records/12608602}.

\bibitem[Glorioso et~al.(2024)Glorioso, Anthony, Tokpanov, Whittington, Pilault, Ibrahim, and Millidge]{glorioso2024zamba}
Paolo Glorioso, Quentin Anthony, Yury Tokpanov, James Whittington, Jonathan Pilault, Adam Ibrahim, and Beren Millidge.
\newblock Zamba: A compact 7b ssm hybrid model.
\newblock \emph{arXiv preprint arXiv:2405.16712}, 2024.

\bibitem[Goldstein et~al.(2025)Goldstein, Alcaide, Lu, and Cheah]{goldstein2025radlads}
Daniel Goldstein, Eric Alcaide, Janna Lu, and Eugene Cheah.
\newblock Radlads: Rapid attention distillation to linear attention decoders at scale.
\newblock \emph{arXiv preprint arXiv:2505.03005}, 2025.

\bibitem[Grattafiori et~al.(2024)Grattafiori, Dubey, Jauhri, Pandey, Kadian, Al-Dahle, Letman, Mathur, Schelten, Vaughan, et~al.]{grattafiori2024llama}
Aaron Grattafiori, Abhimanyu Dubey, Abhinav Jauhri, Abhinav Pandey, Abhishek Kadian, Ahmad Al-Dahle, Aiesha Letman, Akhil Mathur, Alan Schelten, Alex Vaughan, et~al.
\newblock The llama 3 herd of models.
\newblock \emph{arXiv preprint arXiv:2407.21783}, 2024.

\bibitem[Gu and Dao(2023)]{gu2023mamba}
Albert Gu and Tri Dao.
\newblock Mamba: Linear-time sequence modeling with selective state spaces.
\newblock \emph{arXiv preprint arXiv:2312.00752}, 2023.

\bibitem[Gu et~al.(2025)Gu, Hu, Yang, Xi, Chen, Han, and Cai]{gu2025jet}
Yuxian Gu, Qinghao Hu, Shang Yang, Haocheng Xi, Junyu Chen, Song Han, and Han Cai.
\newblock Jet-nemotron: Efficient language model with post neural architecture search.
\newblock \emph{arXiv preprint arXiv:2508.15884}, 2025.

\bibitem[Guo et~al.(2025)Guo, Yang, Zhang, Song, Wang, Zhu, Xu, Zhang, Ma, Bi, et~al.]{guo2025deepseek}
Daya Guo, Dejian Yang, Haowei Zhang, Junxiao Song, Peiyi Wang, Qihao Zhu, Runxin Xu, Ruoyu Zhang, Shirong Ma, Xiao Bi, et~al.
\newblock Deepseek-r1: Incentivizing reasoning capability in llms via reinforcement learning.
\newblock \emph{arXiv preprint arXiv:2501.12948}, 2025.

\bibitem[Hsieh et~al.(2024)Hsieh, Sun, Kriman, Acharya, Rekesh, Jia, Zhang, and Ginsburg]{hsieh2024ruler}
Cheng-Ping Hsieh, Simeng Sun, Samuel Kriman, Shantanu Acharya, Dima Rekesh, Fei Jia, Yang Zhang, and Boris Ginsburg.
\newblock Ruler: What's the real context size of your long-context language models?
\newblock \emph{arXiv preprint arXiv:2404.06654}, 2024.

\bibitem[Hu et~al.(2025)Hu, Pan, Du, Lan, Tang, Wen, Liang, and Sun]{hu2025comba}
Jiaxi Hu, Yongqi Pan, Jusen Du, Disen Lan, Xiaqiang Tang, Qingsong Wen, Yuxuan Liang, and Weigao Sun.
\newblock Comba: Improving bilinear rnns with closed-loop control.
\newblock \emph{arXiv preprint arXiv:2506.02475}, 2025.

\bibitem[Hu et~al.(2024)Hu, Tu, Han, He, Cui, Long, Zheng, Fang, Huang, Zhao, et~al.]{hu2024minicpm}
Shengding Hu, Yuge Tu, Xu~Han, Chaoqun He, Ganqu Cui, Xiang Long, Zhi Zheng, Yewei Fang, Yuxiang Huang, Weilin Zhao, et~al.
\newblock Minicpm: Unveiling the potential of small language models with scalable training strategies.
\newblock \emph{arXiv preprint arXiv:2404.06395}, 2024.

\bibitem[Kasai et~al.(2021)Kasai, Peng, Zhang, Yogatama, Ilharco, Pappas, Mao, Chen, and Smith]{kasai2021finetuning}
Jungo Kasai, Hao Peng, Yizhe Zhang, Dani Yogatama, Gabriel Ilharco, Nikolaos Pappas, Yi~Mao, Weizhu Chen, and Noah~A Smith.
\newblock Finetuning pretrained transformers into rnns.
\newblock In \emph{Proceedings of the 2021 conference on empirical methods in natural language processing}, pages 10630--10643, 2021.

\bibitem[Katharopoulos et~al.(2020)Katharopoulos, Vyas, Pappas, and Fleuret]{katharopoulos2020transformers}
Angelos Katharopoulos, Apoorv Vyas, Nikolaos Pappas, and Fran{\c{c}}ois Fleuret.
\newblock Transformers are rnns: Fast autoregressive transformers with linear attention.
\newblock In \emph{International conference on machine learning}, pages 5156--5165. PMLR, 2020.

\bibitem[Kwon et~al.(2023)Kwon, Li, Zhuang, Sheng, Zheng, Yu, Gonzalez, Zhang, and Stoica]{kwon2023efficient}
Woosuk Kwon, Zhuohan Li, Siyuan Zhuang, Ying Sheng, Lianmin Zheng, Cody~Hao Yu, Joseph Gonzalez, Hao Zhang, and Ion Stoica.
\newblock Efficient memory management for large language model serving with pagedattention.
\newblock In \emph{Proceedings of the 29th symposium on operating systems principles}, pages 611--626, 2023.

\bibitem[Lahoti et~al.(2026)Lahoti, Li, Chen, Wang, Bick, Kolter, Dao, and Gu]{lahoti2026mamba}
Aakash Lahoti, Kevin~Y Li, Berlin Chen, Caitlin Wang, Aviv Bick, J~Zico Kolter, Tri Dao, and Albert Gu.
\newblock Mamba-3: Improved sequence modeling using state space principles.
\newblock \emph{arXiv preprint arXiv:2603.15569}, 2026.

\bibitem[Lan et~al.(2025)Lan, Sun, Hu, Du, and Cheng]{lan2025liger}
Disen Lan, Weigao Sun, Jiaxi Hu, Jusen Du, and Yu~Cheng.
\newblock Liger: Linearizing large language models to gated recurrent structures.
\newblock \emph{arXiv preprint arXiv:2503.01496}, 2025.

\bibitem[Li et~al.(2024)Li, Fang, Smyrnis, Ivgi, Jordan, Gadre, Bansal, Guha, Keh, Arora, et~al.]{li2024datacomp}
Jeffrey Li, Alex Fang, Georgios Smyrnis, Maor Ivgi, Matt Jordan, Samir Gadre, Hritik Bansal, Etash Guha, Sedrick Keh, Kushal Arora, et~al.
\newblock Datacomp-lm: In search of the next generation of training sets for language models.
\newblock \emph{Advances in Neural Information Processing Systems}, 37:\penalty0 14200--14282, 2024.

\bibitem[Li et~al.(2025)Li, Yang, Tan, Mishra, Panda, Zhou, and Kim]{li2025distilling}
Yanhong Li, Songlin Yang, Shawn Tan, Mayank Mishra, Rameswar Panda, Jiawei Zhou, and Yoon Kim.
\newblock Distilling to hybrid attention models via kl-guided layer selection.
\newblock \emph{arXiv preprint arXiv:2512.20569}, 2025.

\bibitem[Lieber et~al.(2024)Lieber, Lenz, Bata, Cohen, Osin, Dalmedigos, Safahi, Meirom, Belinkov, Shalev-Shwartz, et~al.]{lieber2024jamba}
Opher Lieber, Barak Lenz, Hofit Bata, Gal Cohen, Jhonathan Osin, Itay Dalmedigos, Erez Safahi, Shaked Meirom, Yonatan Belinkov, Shai Shalev-Shwartz, et~al.
\newblock Jamba: A hybrid transformer-mamba language model.
\newblock \emph{arXiv preprint arXiv:2403.19887}, 2024.

\bibitem[Lin et~al.(2026)Lin, Li, Chen, Shi, Chen, Hu, and Zhang]{lin2026lycheedecode}
Gang Lin, Dongfang Li, Zhuoen Chen, Yukun Shi, Xuhui Chen, Baotian Hu, and Min Zhang.
\newblock Lycheedecode: Accelerating long-context llm inference via hybrid-head sparse decoding.
\newblock \emph{arXiv preprint arXiv:2602.04541}, 2026.

\bibitem[Liu et~al.(2024)Liu, Feng, Xue, Wang, Wu, Lu, Zhao, Deng, Zhang, Ruan, et~al.]{liu2024deepseek}
Aixin Liu, Bei Feng, Bing Xue, Bingxuan Wang, Bochao Wu, Chengda Lu, Chenggang Zhao, Chengqi Deng, Chenyu Zhang, Chong Ruan, et~al.
\newblock Deepseek-v3 technical report.
\newblock \emph{arXiv preprint arXiv:2412.19437}, 2024.

\bibitem[Lockard et~al.(2019)Lockard, Shiralkar, and Dong]{lockard2019openceres}
Colin Lockard, Prashant Shiralkar, and Xin~Luna Dong.
\newblock Openceres: When open information extraction meets the semi-structured web.
\newblock In \emph{Proceedings of the 2019 Conference of the North American Chapter of the Association for Computational Linguistics: Human Language Technologies, Volume 1 (Long and Short Papers)}, pages 3047--3056, 2019.

\bibitem[Loshchilov and Hutter(2017)]{loshchilov2017decoupled}
Ilya Loshchilov and Frank Hutter.
\newblock Decoupled weight decay regularization.
\newblock \emph{arXiv preprint arXiv:1711.05101}, 2017.

\bibitem[Mercat et~al.(2024)Mercat, Vasiljevic, Keh, Arora, Dave, Gaidon, and Kollar]{mercat2024linearizing}
Jean Mercat, Igor Vasiljevic, Sedrick Keh, Kushal Arora, Achal Dave, Adrien Gaidon, and Thomas Kollar.
\newblock Linearizing large language models.
\newblock \emph{arXiv preprint arXiv:2405.06640}, 2024.

\bibitem[Merrill et~al.(2026)Merrill, Li, Romero, Svete, Costello, Dasigi, Groeneveld, Heineman, Kuehl, Lambert, et~al.]{merrill2026olmo}
William Merrill, Yanhong Li, Tyler Romero, Anej Svete, Caia Costello, Pradeep Dasigi, Dirk Groeneveld, David Heineman, Bailey Kuehl, Nathan Lambert, et~al.
\newblock Olmo hybrid: From theory to practice and back.
\newblock \emph{arXiv preprint arXiv:2604.03444}, 2026.

\bibitem[Paliotta et~al.(2025)Paliotta, Wang, Pagliardini, Li, Bick, Kolter, Gu, Fleuret, and Dao]{paliotta2025thinking}
Daniele Paliotta, Junxiong Wang, Matteo Pagliardini, Kevin~Y Li, Aviv Bick, J~Zico Kolter, Albert Gu, Fran{\c{c}}ois Fleuret, and Tri Dao.
\newblock Thinking slow, fast: Scaling inference compute with distilled reasoners.
\newblock \emph{arXiv preprint arXiv:2502.20339}, 2025.

\bibitem[Peng et~al.(2023)Peng, Alcaide, Anthony, Albalak, Arcadinho, Biderman, Cao, Cheng, Chung, Derczynski, et~al.]{peng2023rwkv}
Bo~Peng, Eric Alcaide, Quentin Anthony, Alon Albalak, Samuel Arcadinho, Stella Biderman, Huanqi Cao, Xin Cheng, Michael Chung, Leon Derczynski, et~al.
\newblock Rwkv: Reinventing rnns for the transformer era.
\newblock In \emph{Findings of the association for computational linguistics: EMNLP 2023}, pages 14048--14077, 2023.

\bibitem[Peng et~al.(2024)Peng, Quesnelle, Fan, and Shippole]{peng2024yarn}
Bowen Peng, Jeffrey Quesnelle, Honglu Fan, and Enrico Shippole.
\newblock Yarn: Efficient context window extension of large language models.
\newblock In \emph{International Conference on Learning Representations}, volume 2024, pages 31932--31951, 2024.

\bibitem[Qin et~al.(2023)Qin, Yang, and Zhong]{qin2023hierarchically}
Zhen Qin, Songlin Yang, and Yiran Zhong.
\newblock Hierarchically gated recurrent neural network for sequence modeling.
\newblock \emph{Advances in Neural Information Processing Systems}, 36:\penalty0 33202--33221, 2023.

\bibitem[Qin et~al.(2024{\natexlab{a}})Qin, Sun, Li, Shen, Sun, and Zhong]{qin2024lightning}
Zhen Qin, Weigao Sun, Dong Li, Xuyang Shen, Weixuan Sun, and Yiran Zhong.
\newblock Lightning attention-2: A free lunch for handling unlimited sequence lengths in large language models.
\newblock \emph{arXiv preprint arXiv:2401.04658}, 2024{\natexlab{a}}.

\bibitem[Qin et~al.(2024{\natexlab{b}})Qin, Yang, Sun, Shen, Li, Sun, and Zhong]{qin2024hgrn2}
Zhen Qin, Songlin Yang, Weixuan Sun, Xuyang Shen, Dong Li, Weigao Sun, and Yiran Zhong.
\newblock Hgrn2: Gated linear rnns with state expansion.
\newblock \emph{arXiv preprint arXiv:2404.07904}, 2024{\natexlab{b}}.

\bibitem[Qiu et~al.(2026)Qiu, Wang, Zheng, Huang, Wen, Yang, Men, Yu, Huang, Huang, et~al.]{qiu2026gated}
Zihan Qiu, Zekun Wang, Bo~Zheng, Zeyu Huang, Kaiyue Wen, Songlin Yang, Rui Men, Le~Yu, Fei Huang, Suozhi Huang, et~al.
\newblock Gated attention for large language models: Non-linearity, sparsity, and attention-sink-free.
\newblock \emph{Advances in Neural Information Processing Systems}, 38:\penalty0 100092--100118, 2026.

\bibitem[{Qwen Team}()]{qwen_qwen3_coder_next_tech_report}
{Qwen Team}.
\newblock Qwen3-coder-next technical report.
\newblock Technical report.
\newblock URL \url{https://github.com/QwenLM/Qwen3-Coder/blob/main/qwen3_coder_next_tech_report.pdf}.
\newblock Accessed: 2026-02-03.

\bibitem[{Qwen Team}(2026)]{qwen3.5}
{Qwen Team}.
\newblock {Qwen3.5}: Towards native multimodal agents, February 2026.
\newblock URL \url{https://qwen.ai/blog?id=qwen3.5}.

\bibitem[Rajpurkar et~al.(2018)Rajpurkar, Jia, and Liang]{rajpurkar2018know}
Pranav Rajpurkar, Robin Jia, and Percy Liang.
\newblock Know what you don’t know: Unanswerable questions for squad.
\newblock In \emph{Proceedings of the 56th Annual Meeting of the Association for Computational Linguistics (Volume 2: Short Papers)}, pages 784--789, 2018.

\bibitem[Ren et~al.(2025)Ren, Liu, Lu, Liang, Chen, et~al.]{ren2025samba}
Liliang Ren, Yang Liu, Yadong Lu, Chen Liang, Weizhu Chen, et~al.
\newblock Samba: Simple hybrid state space models for efficient unlimited context language modeling.
\newblock In \emph{International Conference on Learning Representations}, volume 2025, pages 53551--53575, 2025.

\bibitem[Sakaguchi et~al.(2021)Sakaguchi, Bras, Bhagavatula, and Choi]{sakaguchi2021winogrande}
Keisuke Sakaguchi, Ronan~Le Bras, Chandra Bhagavatula, and Yejin Choi.
\newblock Winogrande: An adversarial winograd schema challenge at scale.
\newblock \emph{Communications of the ACM}, 64\penalty0 (9):\penalty0 99--106, 2021.

\bibitem[Sun et~al.(2025{\natexlab{a}})Sun, Hu, Zhou, Du, Lan, Wang, Zhu, Qu, Zhang, Mo, et~al.]{sun2025speed}
Weigao Sun, Jiaxi Hu, Yucheng Zhou, Jusen Du, Disen Lan, Kexin Wang, Tong Zhu, Xiaoye Qu, Yu~Zhang, Xiaoyu Mo, et~al.
\newblock Speed always wins: A survey on efficient architectures for large language models.
\newblock \emph{arXiv preprint arXiv:2508.09834}, 2025{\natexlab{a}}.

\bibitem[Sun et~al.(2025{\natexlab{b}})Sun, Lan, Zhu, Qu, and Cheng]{sun2025linear}
Weigao Sun, Disen Lan, Tong Zhu, Xiaoye Qu, and Yu~Cheng.
\newblock Linear-moe: Linear sequence modeling meets mixture-of-experts.
\newblock \emph{arXiv preprint arXiv:2503.05447}, 2025{\natexlab{b}}.

\bibitem[Team et~al.(2025)Team, Zhang, Lin, Yao, Hu, Meng, Liu, Men, Yang, Li, et~al.]{team2025kimi_linear}
Kimi Team, Yu~Zhang, Zongyu Lin, Xingcheng Yao, Jiaxi Hu, Fanqing Meng, Chengyin Liu, Xin Men, Songlin Yang, Zhiyuan Li, et~al.
\newblock Kimi linear: An expressive, efficient attention architecture.
\newblock \emph{arXiv preprint arXiv:2510.26692}, 2025.

\bibitem[Vaswani et~al.(2017)Vaswani, Shazeer, Parmar, Uszkoreit, Jones, Gomez, Kaiser, and Polosukhin]{vaswani2017attention}
Ashish Vaswani, Noam Shazeer, Niki Parmar, Jakob Uszkoreit, Llion Jones, Aidan~N Gomez, {\L}ukasz Kaiser, and Illia Polosukhin.
\newblock Attention is all you need.
\newblock \emph{Advances in neural information processing systems}, 30, 2017.

\bibitem[Wang et~al.(2025)Wang, Zhu, Abreu, Shan, Kergan, Pan, Chou, Li, Zhang, Huang, et~al.]{wang2025systematic}
Dustin Wang, Rui-Jie Zhu, Steven Abreu, Yong Shan, Taylor Kergan, Yuqi Pan, Yuhong Chou, Zheng Li, Ge~Zhang, Wenhao Huang, et~al.
\newblock A systematic analysis of hybrid linear attention.
\newblock \emph{arXiv preprint arXiv:2507.06457}, 2025.

\bibitem[Wang et~al.(2024)Wang, Paliotta, May, Rush, and Dao]{wang2024mamba}
Junxiong Wang, Daniele Paliotta, Avner May, Alexander~M Rush, and Tri Dao.
\newblock The mamba in the llama: Distilling and accelerating hybrid models.
\newblock \emph{Advances in Neural Information Processing Systems}, 37:\penalty0 62432--62457, 2024.

\bibitem[Wen et~al.(2025)Wen, Dang, and Lyu]{wen2025rnns}
Kaiyue Wen, Xingyu Dang, and Kaifeng Lyu.
\newblock Rnns are not transformers (yet): The key bottleneck on in-context retrieval.
\newblock In \emph{International Conference on Learning Representations}, volume 2025, pages 48813--48856, 2025.

\bibitem[Xiao et~al.(2025)Xiao, Tang, Zuo, Guo, Yang, Tang, Fu, and Han]{xiao2025duoattention}
Guangxuan Xiao, Jiaming Tang, Jingwei Zuo, Junxian Guo, Shang Yang, Haotian Tang, Yao Fu, and Song Han.
\newblock Duoattention: Efficient long-context llm inference with retrieval and streaming heads.
\newblock In \emph{International Conference on Learning Representations}, volume 2025, pages 37228--37253, 2025.

\bibitem[Yang et~al.(2025)Yang, Li, Yang, Zhang, Hui, Zheng, Yu, Gao, Huang, Lv, et~al.]{yang2025qwen3}
An~Yang, Anfeng Li, Baosong Yang, Beichen Zhang, Binyuan Hui, Bo~Zheng, Bowen Yu, Chang Gao, Chengen Huang, Chenxu Lv, et~al.
\newblock Qwen3 technical report.
\newblock \emph{arXiv preprint arXiv:2505.09388}, 2025.

\bibitem[Yang et~al.(2026)Yang, Rezagholizadeh, Li, Appia, and Barsoum]{yang2026zebra}
Mingyu Yang, Mehdi Rezagholizadeh, Guihong Li, Vikram Appia, and Emad Barsoum.
\newblock Zebra-llama: Towards extremely efficient hybrid models.
\newblock \emph{Advances in Neural Information Processing Systems}, 38:\penalty0 78167--78194, 2026.

\bibitem[Yang and Zhang(2024)]{yang2024fla}
Songlin Yang and Yu~Zhang.
\newblock Fla: A triton-based library for hardware-efficient implementations of linear attention mechanism, January 2024.
\newblock URL \url{https://github.com/fla-org/flash-linear-attention}.

\bibitem[Yang et~al.(2023)Yang, Wang, Shen, Panda, and Kim]{yang2023gated}
Songlin Yang, Bailin Wang, Yikang Shen, Rameswar Panda, and Yoon Kim.
\newblock Gated linear attention transformers with hardware-efficient training.
\newblock \emph{arXiv preprint arXiv:2312.06635}, 2023.

\bibitem[Yang et~al.(2024{\natexlab{a}})Yang, Kautz, and Hatamizadeh]{yang2024gated}
Songlin Yang, Jan Kautz, and Ali Hatamizadeh.
\newblock Gated delta networks: Improving mamba2 with delta rule.
\newblock \emph{arXiv preprint arXiv:2412.06464}, 2024{\natexlab{a}}.

\bibitem[Yang et~al.(2024{\natexlab{b}})Yang, Wang, Zhang, Shen, and Kim]{yang2024parallelizing}
Songlin Yang, Bailin Wang, Yu~Zhang, Yikang Shen, and Yoon Kim.
\newblock Parallelizing linear transformers with the delta rule over sequence length.
\newblock \emph{Advances in neural information processing systems}, 37:\penalty0 115491--115522, 2024{\natexlab{b}}.

\bibitem[Zellers et~al.(2019)Zellers, Holtzman, Bisk, Farhadi, and Choi]{zellers2019hellaswag}
Rowan Zellers, Ari Holtzman, Yonatan Bisk, Ali Farhadi, and Yejin Choi.
\newblock Hellaswag: Can a machine really finish your sentence?
\newblock In \emph{Proceedings of the 57th annual meeting of the association for computational linguistics}, pages 4791--4800, 2019.

\bibitem[Zhang et~al.(2024{\natexlab{a}})Zhang, Arora, Chalamala, Wu, Spector, Singhal, Ramesh, and R{\'e}]{zhang2024lolcats}
Michael Zhang, Simran Arora, Rahul Chalamala, Alan Wu, Benjamin Spector, Aaryan Singhal, Krithik Ramesh, and Christopher R{\'e}.
\newblock Lolcats: On low-rank linearizing of large language models.
\newblock \emph{arXiv preprint arXiv:2410.10254}, 2024{\natexlab{a}}.

\bibitem[Zhang et~al.(2024{\natexlab{b}})Zhang, Bhatia, Kumbong, and R{\'e}]{zhang2024hedgehog}
Michael Zhang, Kush Bhatia, Hermann Kumbong, and Christopher R{\'e}.
\newblock The hedgehog \& the porcupine: Expressive linear attentions with softmax mimicry.
\newblock \emph{arXiv preprint arXiv:2402.04347}, 2024{\natexlab{b}}.

\bibitem[Zhang et~al.(2024{\natexlab{c}})Zhang, Yang, Zhu, Zhang, Cui, Wang, Wang, Shi, Wang, Bi, et~al.]{zhang2024gated}
Yu~Zhang, Songlin Yang, Ruijie Zhu, Yue Zhang, Leyang Cui, Yiqiao Wang, Bolun Wang, Freda Shi, Bailin Wang, Wei Bi, et~al.
\newblock Gated slot attention for efficient linear-time sequence modeling.
\newblock \emph{Advances in Neural Information Processing Systems}, 37:\penalty0 116870--116898, 2024{\natexlab{c}}.

\bibitem[Zuo et~al.(2025)Zuo, Velikanov, Chahed, Belkada, Rhayem, Kunsch, Hacid, Yous, Farhat, Khadraoui, et~al.]{zuo2025falcon}
Jingwei Zuo, Maksim Velikanov, Ilyas Chahed, Younes Belkada, Dhia~Eddine Rhayem, Guillaume Kunsch, Hakim Hacid, Hamza Yous, Brahim Farhat, Ibrahim Khadraoui, et~al.
\newblock Falcon-h1: A family of hybrid-head language models redefining efficiency and performance.
\newblock \emph{arXiv preprint arXiv:2507.22448}, 2025.

\end{thebibliography}

\clearpage

\beginappendix

\section{Model and Training Configuration}
\label{app:morph_config}

We report the complete model and training configuration in Table~\ref{tab:morph_config}. All training data are randomly sampled from the DCLM corpus~\citep{li2024datacomp}; in each stage, we use only a token subset of the specified size to control the data budget across model variants and training stages. We use the AdamW optimizer~\citep{loshchilov2017decoupled} with beta values of $(0.9, 0.95)$ and a weight decay of $0.0$ throughout all training stages. Following HALO~\citep{chen2026hybrid}, the overall Transformer-to-hybrid conversion training pipeline of FlashMorph consists of four stages: hidden-state alignment for morphable layer construction, layer selection, distillation, and long-context finetuning. Stage-specific hyperparameters are summarized in Table~\ref{tab:morph_config}.

\section{Implementation Details}
\label{app:imp}

During the layer selection stage, we construct a synthetic long-context retrieval dataset based on the DCLM corpus~\citep{li2024datacomp}. For each example, we use DCLM text as the background context and insert ten randomly generated passkey sequences at different depths. Each passkey contains $s$ words sampled from a fixed alphabet, with $s=32$ in our experiments. The context length is randomly sampled from 50 length intervals ranging from 1K to 16K tokens. For insertion, we discretize the context depth into 1,000 candidate points and randomly sample the insertion positions of the passkeys. The model is then required to recall all ten passkeys at the end of the context. This design provides dense long-range retrieval supervision, enabling us to identify which layers can be replaced by linear attention while preserving the model's ability to recover information from distant positions under long-context scenarios.

\begin{tcolorbox}[
    enhanced,
    breakable,
    colframe=blue!90!black,
    colback=blue!10!white,
    coltitle=white,
    fonttitle=\bfseries,
    title=Case Sample from Synthetic Retrieval Data,
    sharp corners,
    boxrule=0.5mm,
    before upper={\small\ttfamily\raggedright\sloppy}
]
\label{app:case}
{\normalfont\bfseries query}

\medskip
<|im\_start|> This is a very long story book: <book>

\medskip
{\color{gray!70!black}... background DCLM text ...}

\medskip
Remember this sequence of words, it is the first passkey to the vault:

\PK{Passkey 1}{xray whiskey lima charlie papa hotel mike india quebec mike victor echo kilo
lima charlie india yankee golf golf delta uniform alpha whiskey golf yankee
zulu charlie november november charlie charlie golf}

\medskip
{\color{gray!70!black}... more DCLM text with the second to ninth passkeys inserted
at different depths ...}

\medskip
Remember this sequence of words, it is the tenth passkey to the vault:

\PK{Passkey 10}{whiskey hotel charlie sierra xray kilo golf uniform alpha bravo alpha lima
november quebec november victor xray whiskey tango juliett papa india delta
uniform alpha mike golf uniform india delta quebec zulu}

\medskip
</book>.

\medskip
{\normalfont\bfseries response}

\medskip
Based on the content of the book, what is the first passkey to the vault?

\PK{Answer 1}{Passkey: xray whiskey lima charlie papa hotel mike india quebec mike
victor echo kilo lima charlie india yankee golf golf delta uniform alpha
whiskey golf yankee zulu charlie november november charlie charlie golf}

\medskip
{\color{gray!70!black}...}

\medskip
Based on the content of the book, what is the tenth passkey to the vault?

\PK{Answer 10}{Passkey: whiskey hotel charlie sierra xray kilo golf uniform alpha bravo
alpha lima november quebec november victor xray whiskey tango juliett papa india
delta uniform alpha mike golf uniform india delta quebec zulu}
\end{tcolorbox}

\begin{table*}[!htbp]
\centering
\caption{\textbf{Hyperparameter Configurations for FlashMorph Models and Training Pipeline.}}
\label{tab:morph_config}
\small
\begin{tabular}{llcccc}
\toprule
\textbf{Setting} & \textbf{Hyperparameter} & \textbf{0.6B} & \textbf{1.7B} & \textbf{8B} & \textbf{30B-A3B} \\
\midrule
\multicolumn{6}{c}{\textit{Model Architecture}} \\
\midrule
\multirow{10}{*}{\textbf{Backbone}}
& \#layers & 28 & 28 & 36  & 48  \\
& hidden size & 1024 & 2048 & 4096 & 2048 \\
& FFN width & 3072 & 6144 & 12288 & 6144 \\
& \#full-attention layers & 7 & 7 & 9  & 12\\
& \#linear-attention layers & 21 & 21 & 27 & 36 \\
& head dimension & 128 & 128 & 128 & 128 \\
& \#attention heads & 16 & 16 & 32 & 32 \\
& \#full-attention KV heads & 8 & 8 & 8 & 4 \\
& \#linear-attention KV heads & 16 & 16 & 32 & 32 \\
& attn. logits scaling $\alpha$ \textit{(if applicable)} & 300 & 500 & 900 & 600 \\

\midrule
\multicolumn{6}{c}{\textit{Training Pipeline}} \\
\midrule
\multirow{7}{*}{\shortstack[l]{\textbf{Stage 1}\\(Hidden-state Alignment)}}
& tokens & \multicolumn{4}{c}{320M} \\
& LR scheduler & \multicolumn{4}{c}{cosine} \\
& learning rate & \multicolumn{4}{c}{$1\mathrm{e}{-3} \rightarrow 1\mathrm{e}{-5}$} \\
& sequence length & \multicolumn{4}{c}{512} \\
& batch size & \multicolumn{4}{c}{32} \\
& warmup steps & \multicolumn{4}{c}{50} \\
& training steps & \multicolumn{4}{c}{20,000} \\

\midrule
\multirow{8}{*}{\shortstack[l]{\textbf{Stage 2}\\(Layer Selection)}}
& tokens & \multicolumn{4}{c}{20M} \\
& LR scheduler & \multicolumn{4}{c}{WSD~\citep{hu2024minicpm}} \\
& learning rate & \multicolumn{4}{c}{$2\mathrm{e}{-2} \rightarrow 2\mathrm{e}{-3}$} \\
& sequence length & \multicolumn{4}{c}{$<16$K} \\
& batch size & \multicolumn{4}{c}{8} \\
& warmup steps & \multicolumn{4}{c}{50} \\
& decay steps & \multicolumn{4}{c}{50} \\
& training steps & \multicolumn{4}{c}{250} \\

\midrule
\multirow{6}{*}{\shortstack[l]{\textbf{Stage 3}\\(Distillation)}}
& tokens & \multicolumn{4}{c}{1B} \\
& LR scheduler & \multicolumn{4}{c}{cosine} \\
& learning rate & \multicolumn{4}{c}{$1\mathrm{e}{-4} \rightarrow 1\mathrm{e}{-5}$} \\
& sequence length & \multicolumn{4}{c}{512} \\
& batch size & \multicolumn{4}{c}{96} \\
& warmup steps & \multicolumn{4}{c}{50} \\
& training steps & \multicolumn{4}{c}{20,000} \\
\midrule
\multirow{6}{*}{\shortstack[l]{\textbf{Stage 4}\\(Long-context Finetuning)}}
& tokens & \multicolumn{4}{c}{1B} \\
& LR scheduler & \multicolumn{4}{c}{constant} \\
& learning rate & \multicolumn{4}{c}{$1\mathrm{e}{-5}$} \\
& sequence length & \multicolumn{4}{c}{16K} \\
& batch size & \multicolumn{4}{c}{128} \\
& warmup steps & \multicolumn{4}{c}{50} \\
& training steps & \multicolumn{4}{c}{500} \\

\bottomrule
\end{tabular}
\end{table*}

\begin{table*}[t]
\centering
\caption{\textbf{NIAH Performance on 8B Dense and 30B-A3B MoE Backbones across 32K-256K Context Lengths.} The best and second-best results are marked sin \textbf{bold} and \underline{underlined}, respectively.}
\vspace{-0.5em}
\small
\resizebox{0.97\linewidth}{!}{
\begin{tabular}{lc|cccc|cccc|cccc}
    \toprule
    \multirow{2}{*}{\textbf{Model}} & \multirow{2}{*}{\textbf{\shortstack{LS. Tokens}}} & \multicolumn{4}{c}{\textbf{NIAH-Single-1}} & \multicolumn{4}{c}{\textbf{NIAH-Single-2}} & \multicolumn{4}{c}{\textbf{NIAH-Single-3}} \\
    \cmidrule(lr){3-14} 
    & & 32K & 64K & 128K & 256K & 32K & 64K & 128K & 256K & 32K & 64K & 128K & 256K  \\
    \midrule
    \multicolumn{14}{l}{\textit{8B backbone}} \\
    \rowcolor{gray!15} 
    Qwen3  & - & 100 & 100 & 52.8 & 0 & 100 & 100 & 0 & 0 & 100 & 99.8 & 0 & 0 \\
    \rowcolor{gray!15} 
    Qwen3+YaRN & - & 100 & 100 & 100 & 100 & 99.4 & 99.6 & 98.6 & 74.8 & 99.6 & 98.0 & 98.8 & 92.8   \\
    \midrule
    Uniform & N/A & 91.8 & 92.4 & 93.2 & 92.6 & 91.4 & 60.6 & 31.6 & 17.8 & 58.2 & 54.2& 40.6 & 26.2 \\
    HALO  & 234M &  \underline{99.2} & \underline{99.4} & \underline{99.2} & \underline{98.4} & \underline{96.8} & \underline{95.6} & \textbf{89.2} & \underline{68.4} & \underline{89.8} & \underline{85.2} & \underline{75.0} & \underline{50.8} \\
    \rowcolor{cyan!20} 
    \textbf{FlashMorph}  & \textbf{20M} & \textbf{99.8} & \textbf{99.6} & \textbf{99.6} & \textbf{99.2} & \textbf{98.0} & \textbf{99.4} & \underline{85.6} & \textbf{82.6} & \textbf{99.4} & \textbf{98.2} & \textbf{92.4} & \textbf{94.0} \\
    \midrule
    \multicolumn{14}{l}{\textit{30B-A3B backbone}} \\
    \rowcolor{gray!15} 
    Qwen3  & - & 100 & 100 & 1.0 & 0  & 100 & 99.8 & 0 & 0 & 100 & 100 & 0 & 0 \\
    \rowcolor{gray!15} 
    Qwen3+YaRN & - & 98.2 & 4.4 & 2.0 & 20.0 & 84.6 & 61.0 & 79.6 & 73.0 & 37.4 & 17.2 & 26.6 & 9.0 \\
    \midrule
    Uniform & N/A & \underline{98.4} & \textbf{99.4} & \underline{99.0} & \textbf{99.6} & \underline{73.6} & 53.4 & \underline{35.4} & \underline{19.0} & \textbf{55.2} & \underline{36.0} & \underline{18.4} & \underline{9.6} \\
    HALO  & 234M & 97.6 & \underline{98.6} & \textbf{99.4} & \underline{99.0} & \textbf{94.8} & \textbf{79.4} & \textbf{61.2} & \textbf{27.6} & \underline{47.0} & 17.8 & 12.0 & 0.8  \\
    \rowcolor{cyan!20} 
    \textbf{FlashMorph}  & \textbf{20M} & \textbf{98.6} & 96.2 & 94.0 & 91.4 & 70.2 & \underline{53.6} & 21.4 & 9.6 &  32.6 & \textbf{38.2} & \textbf{24.6} & \textbf{18.2} \\
    \bottomrule
\end{tabular}
}
\addtolength{\tabcolsep}{2.5pt}    
\centering
\label{tab:niah2}
\end{table*}

\vspace{-0.5em}
\begin{table*}[!htbp]
\centering
\caption{\textbf{Zero-shot Performance on Commonsense Reasoning and Long-context Recall-intensive Tasks under the HypeNet Setting.} The best and second-best results are marked in \textbf{bold} and \underline{underlined}, respectively.}
\label{tab:zero_shot2}
\vspace{-0.5em}
\resizebox{0.97\textwidth}{!}{
\begin{tabular}{lc|cccccc|cccc}
\toprule
\multirow{2}{*}{\textbf{Method}}
& \multirow{2}{*}{\textbf{LS. Tokens}}
& \textbf{ARC-e} & \textbf{ARC-c} & \textbf{PIQA} & \textbf{Hella.} & \textbf{Wino.}
& \multirow{2}{*}{\textbf{Avg.}}
& \textbf{SQuAD} & \textbf{FDA} & \textbf{SWDE} & \multirow{2}{*}{\textbf{Avg.}} \\
& & acc & acc$_n$ & acc & acc$_n$ & acc & & acc & acc & acc & \\
\midrule
\multicolumn{12}{l}{\textit{0.6B backbone}} \\
\rowcolor{gray!15}
Qwen3 & - &  60.6 & 34.1 & 67.6 & 47.3 & 55.7 & 53.1 & 44.1 & 82.1 & 80.5 & 68.9 \\
\midrule
Uniform & N/A  & 61.2 & \textbf{33.1} & 66.8 & 44.7 & \textbf{56.6} & 52.1 & 20.7 & 52.6 & 70.6 & 48.0 \\
KL-LS & 20B & \underline{61.9} & \textbf{33.1} & \textbf{67.8} & \underline{46.4} & 53.9 & \underline{52.6} & \underline{25.7} & \underline{58.4} & 70.8 & 51.6 \\
HALO  & 234M  & \textbf{62.9} & \underline{32.5} & \underline{67.1} & \textbf{46.5} & \underline{55.7} & \textbf{52.9} & 22.8 & \textbf{62.8} & \underline{73.5} & \underline{53.0}  \\
\rowcolor{cyan!15}
\textbf{FlashMorph}  & \textbf{20M}  & \textbf{62.9} & 30.6 & 67.0 & 46.0 & 55.1 & 52.3 & \textbf{35.3 }& 58.2 & \textbf{76.8} & \textbf{56.7}  \\
\midrule
\multicolumn{12}{l}{\textit{1.7B backbone}} \\
\rowcolor{gray!15}
Qwen3 & - & 72.4 & 43.5 & 72.5 & 60.4 & 61.0 & 62.0 & 39.8 & 79.0 & 85.1 & 67.9 \\
\midrule
Uniform & N/A  &  \underline{73.2} & 42.6 & \textbf{73.0} & \underline{60.1} & \underline{62.8} & \textbf{62.4} & \underline{40.2} & 64.5 & 80.4 & 61.7  \\
PostNAS* & 50B & \textbf{73.7} & 42.3 & \underline{72.9} & \textbf{60.3} & 61.6 & 62.1 & \textbf{43.5} & 63.5 & \textbf{81.5}& \underline{62.8} \\
KL-LS & 20B &  72.7 & \underline{42.8} & 72.4 & 59.6 & 58.8 & 61.3 &  26.7 & 48.8 & 73.9 & 49.8 \\
HALO  & 234M  & 72.5 & 41.9 & 72.6 & 60.0 & \textbf{63.7} & 62.1 & 38.5 & \underline{67.8} & 80.7 & 62.3 \\
\rowcolor{cyan!15}
\textbf{FlashMorph}  & \textbf{20M}  & 73.1 & \textbf{43.3} & \textbf{73.0} & \underline{60.1} & 61.8 & \underline{62.3} & 39.8 & \textbf{70.2} & \underline{81.4} & \textbf{63.8} \\
\midrule
\multicolumn{12}{l}{\textit{8B backbone}} \\
\rowcolor{gray!15}
Qwen3 & - & 83.6 & 56.6 & 76.8 & 75.0 & 67.7 & 71.9 & 72.3 & 78.2 & 90.8 & 80.4 \\
\midrule
Uniform & N/A  & 81.4 & 56.5 & \textbf{77.4} & \textbf{73.7} & \underline{70.1} & 71.8 & \underline{49.7} & \underline{63.3} & \underline{84.1} & \underline{65.7}  \\
HALO  & 234M  &  \textbf{82.5} & \underline{57.4} & \underline{77.2} & 72.7 & 69.9 & \underline{71.9} & 41.9 & 59.7 & 80.9 & 60.8 \\
\rowcolor{cyan!15}
\textbf{FlashMorph}  & \textbf{20M}  & \underline{81.9} & \textbf{57.8} & \underline{77.2} & \underline{73.1} & \textbf{71.1} & \textbf{72.2} & \textbf{52.7} & \textbf{73.5} & \textbf{87.3} & \textbf{71.2} \\
\midrule
\multicolumn{12}{l}{\textit{30B-A3B backbone}} \\
\rowcolor{gray!15}
Qwen3 & - & 79.5 & 56.1 & 79.5 & 77.7 & 70.9 & 72.7 & 58.7 & 81.0 & 90.8 & 76.8  \\
\midrule
Uniform & N/A  & \underline{79.3} & \underline{51.9} & \underline{77.9} & \underline{73.1} & \underline{66.5} & \underline{69.8} & \textbf{23.0} & 35.9 & 76.8 & 45.2  \\
HALO  & 234M  &  75.6 & 47.9 & 76.4 & 68.8 & 61.6 & 66.0 & 20.8 & \underline{41.0} & \underline{77.0} & \underline{46.3} \\
\rowcolor{cyan!15}
\textbf{FlashMorph}  & \textbf{20M}  &  \textbf{80.9} & \textbf{56.1} & \textbf{79.9} & \textbf{75.0} & \textbf{72.2} & \textbf{72.8} & \underline{21.4} & \textbf{53.4} & \textbf{81.5} & \textbf{52.1} \\
\bottomrule
\end{tabular}
}
\end{table*}

\vspace{-3.0em}
\section{More Experiment Results}

As presented in Table~\ref{tab:niah2} and Table~\ref{tab:zero_shot2}, we further evaluate FlashMorph on Qwen3-8B and Qwen3-30B-A3B under the HypeNet setting~\citep{chen2026hybrid}, comparing it against uniform interleaving and HALO. We exclude PostNAS and KL-LS at these scales because reproducing their layer-selection procedures would incur prohibitive computational costs, particularly for larger models.

\begin{table}[!ht]
\centering
\caption{\textbf{Complete Layer Selection Results for FlashMorph and Baseline Methods.} Layer indices are sorted from most important to least important, and the boxed prefix denotes the top-25\% layers retained as full-attention layers under the fixed hybrid ratio budget. Red indices in baseline rows indicate selected layers that are not shared with FlashMorph. $^{*}$ denotes the selected layer results taken from~\citep{chen2026hybrid}. }
\label{tab:layer_selection_results}
\resizebox{\linewidth}{!}{%
\begin{tabular}{ll|l}
\toprule
\textbf{Attn.} & \textbf{Method} & \textbf{Layer indices (most important $\rightarrow$ least important)} \\
\midrule
\multicolumn{3}{l}{\textit{Qwem3-0.6B backbone}} \\
\midrule
\multirow{3}{*}{Lightning}
& FlashMorph (Ours) & \fbox{1, 16, 21, 11, 19, 24, 0,} 25, 18, 2, 6, 8, 20, 3, 26, 13, 9, 23, 22, 10, 14, 17, 4, 15, 12, 27, 5, 7 \\
& HALO & \fbox{\textcolor{red}{10}, 21, \textcolor{red}{9}, \textcolor{red}{5}, 11, 1, \textcolor{red}{13},} 16, 25, 12, 19, 24, 6, 18, 15, 8, 2, 26, 14, 23, 0, 27, 20, 7, 17, 22, 3, 4 \\
& KL-LS & \fbox{21, 16, 19, \textcolor{red}{20}, \textcolor{red}{18}, \textcolor{red}{22}, \textcolor{red}{25},} 24, 26, 17, 23, 14, 8, 12, 6, 13, 11, 15, 3, 2, 9, 4, 1, 5, 0, 10, 7, 27 \\
\midrule
\multirow{3}{*}{GLA}
& FlashMorph (Ours) & \fbox{21, 16, 11, 1, 2, 6, 25,} 19, 24, 18, 20, 8, 0, 26, 13, 22, 3, 23, 9, 10, 14, 17, 12, 15, 4, 7, 27, 5 \\
& HALO & \fbox{\textcolor{red}{8}, 21, \textcolor{red}{13}, 1, \textcolor{red}{24}, 6, 25,} 19, 18, 11, 10, 16, 9, 12, 5, 15, 2, 26, 27, 23, 20, 4, 0, 14, 3, 22, 7, 17 \\
& KL-LS & \fbox{21, 16, \textcolor{red}{20}, \textcolor{red}{19}, \textcolor{red}{18}, \textcolor{red}{24}, \textcolor{red}{8},} 22, 17, 6, 23, 25, 26, 14, 12, 11, 13, 2, 3, 15, 9, 4, 1, 0, 5, 10, 7, 27  \\
\midrule
\multirow{3}{*}{GDN}
& FlashMorph (Ours) &  \fbox{1, 11, 21, 16, 19, 18, 24,} 6, 25, 8, 2, 20, 0, 13, 26, 3, 9, 23, 14, 22, 10, 12, 17, 15, 4, 7, 5, 27 \\
& HALO & \fbox{\textcolor{red}{10}, \textcolor{red}{5}, 21, \textcolor{red}{6}, 19, 24, 18,} 11, 9, 13, 12, 25, 8, 2, 16, 26, 1, 4, 15, 0, 17, 7, 23, 14, 27, 3, 22, 20 \\
& KL-LS & \fbox{21, 16, 19, \textcolor{red}{20}, 18, \textcolor{red}{6}, \textcolor{red}{22},} 24, 25, 11, 8, 14, 26, 12, 23, 17, 13, 3, 2, 9, 1, 4, 15, 5, 0, 10, 27, 7  \\
\midrule
\midrule
\multicolumn{3}{l}{\textit{Qwem3-1.7B backbone}} \\
\midrule
\multirow{4}{*}{Lightning}
& FlashMorph (Ours) & \fbox{1, 16, 13, 11, 21, 3, 8,} 20, 6, 19, 18, 9, 14, 24, 2, 17, 0, 10, 15, 25, 23, 26, 22, 12, 4, 5, 7, 27 \\
& HALO & \fbox{3, \textcolor{red}{14}, \textcolor{red}{9}, \textcolor{red}{2}, \textcolor{red}{6}, 16, 21,} 25, 24, 8, 23, 12, 11, 26, 27, 19, 18, 17, 7, 15, 4, 13, 1, 20, 10, 22, 0, 5 \\
& KL-LS &  \fbox{21, 16, \textcolor{red}{20}, \textcolor{red}{19}, \textcolor{red}{18}, \textcolor{red}{22}, \textcolor{red}{24},} 26, 25, 23, 17, 6, 14, 8, 12, 13, 11, 3, 15, 2, 9, 1, 4, 0, 5, 10, 7, 27 \\
& PostNAS* &  \fbox{\textcolor{red}{0}, 21, \textcolor{red}{25}, \textcolor{red}{19}, \textcolor{red}{6}, 11, \textcolor{red}{9},} 24, 12, 2, 26, 16, 17, 23, 18, 4, 7, 3, 14, 20, 1, 27, 10, 13, 8, 22, 15, 5 \\
\midrule
\multirow{4}{*}{GLA}
& FlashMorph (Ours) & \fbox{1, 11, 21, 3, 13, 14, 16,} 19, 9, 20, 6, 18, 2, 0, 24, 8, 10, 17, 25, 15, 23, 26, 12, 22, 4, 7, 27, 5  \\
& HALO &  \fbox{3, 14, \textcolor{red}{6}, \textcolor{red}{8}, \textcolor{red}{4}, \textcolor{red}{25}, 11,} 21, 16, 2, 24, 18, 26, 17, 19, 23, 1, 27, 12, 0, 15, 13, 7, 20, 9, 22, 10, 5 \\
& KL-LS &  \fbox{21, 16, \textcolor{red}{20}, \textcolor{red}{19}, \textcolor{red}{18}, \textcolor{red}{24}, \textcolor{red}{22},} 6, 8, 26, 11, 25, 23, 14, 12, 17, 3, 13, 2, 1, 9, 4, 15, 0, 5, 10, 27, 7 \\
& PostNAS* &  \fbox{\textcolor{red}{0}, 21, \textcolor{red}{25}, \textcolor{red}{19}, \textcolor{red}{6}, 11, \textcolor{red}{9},} 24, 12, 2, 26, 16, 17, 23, 18, 4, 7, 3, 14, 20, 1, 27, 10, 13, 8, 22, 15, 5 \\
\midrule
\multirow{4}{*}{GDN}
& FlashMorph (Ours) &  \fbox{1, 11, 13, 21, 16, 14, 6,} 10, 20, 19, 2, 18, 8, 24, 0, 3, 9, 25, 15, 22, 17, 26, 23, 12, 4, 27, 7, 5 \\
& HALO &  \fbox{\textcolor{red}{3}, \textcolor{red}{2}, \textcolor{red}{25}, 21, 11, 14, 6,} 4, 12, 8, 16, 18, 24, 17, 19, 23, 7, 26, 27, 9, 1, 20, 22, 13, 0, 15, 5, 10 \\
& KL-LS &  \fbox{21, 16, \textcolor{red}{20}, \textcolor{red}{19}, 6, 11, \textcolor{red}{18},} 24, 22, 8, 25, 26, 12, 14, 23, 13, 2, 3, 17, 1, 9, 4, 15, 5, 10, 0, 27, 7 \\
& PostNAS* &  \fbox{\textcolor{red}{0}, 21, \textcolor{red}{25}, \textcolor{red}{19}, 6, 11, \textcolor{red}{9},} 24, 12, 2, 26, 16, 17, 23, 18, 4, 7, 3, 14, 20, 1, 27, 10, 13, 8, 22, 15, 5 \\
\midrule
\midrule
\multicolumn{3}{l}{\textit{Qwen3-8B backbone}} \\
\midrule
\multirow{3}{*}{Lightning}
& FlashMorph (Ours) & \makecell[l]{\fbox{1, 0, 7, 22, 29, 13, 3, 15, 24,} 20, 9, 19, 33, 34, 12, 2, 17, 8, 32, 16, 14, 23, 21, 35, 18, 5, 31, \\11, 10, 4, 28, 26, 30, 6, 25, 27} \\
& HALO & \makecell[l]{\fbox{7, \textcolor{red}{6}, 0, 24, \textcolor{red}{8}, \textcolor{red}{33}, 1, \textcolor{red}{12}, \textcolor{red}{34},} 22, 2, 15, 26, 9, 20, 16, 29, 31, 21, 4, 17, 30, 5, 35, 11, 32, 3, 25, \\27, 14, 18, 19, 28, 13, 23, 10} \\
\midrule
\midrule
\multicolumn{3}{l}{\textit{Qwen3-30B-A3B backbone}} \\
\midrule
\multirow{3}{*}{Lightning}
& FlashMorph (Ours) & \makecell[l]{\fbox{37, 2, 4, 3, 1, 38, 5, 0, 6, 42, 8, 26,} 13, 22, 18, 21, 11, 9, 14, 20, 15, 25, 36, 24, 10, 41, 12, 34,\\ 47, 17, 23, 30, 7, 45, 39, 16, 19, 29, 35, 43, 33, 46, 27, 44, 40, 32, 28, 31} \\
& HALO & \makecell[l]{\fbox{5, \textcolor{red}{45}, 42, 8, \textcolor{red}{12}, \textcolor{red}{43}, \textcolor{red}{36}, \textcolor{red}{19}, \textcolor{red}{34}, \textcolor{red}{41}, \textcolor{red}{18}, \textcolor{red}{22},} 46, 27, 37, 4, 31, 39, 13, 9, 35, 10, 24, 7, 33, 25,47,\\23, 28, 17, 21, 11, 14, 0, 30, 15, 20, 26, 40, 3, 1, 6, 32, 38, 2, 29, 16, 44 } \\
\bottomrule
\end{tabular}
}
\end{table}

\section{Complete Layer Importance Ranking}

We provide the complete layer importance rankings in Table~\ref{tab:layer_selection_results}. For each backbone and linear-attention variant, layers are sorted from the most to the least important, and the top-ranked layers are retained as full-attention layers under the fixed hybrid budget.

\end{document}